\documentclass[pdflatex,sn-mathphys-num]{sn-jnl}

\usepackage{graphicx}%
\usepackage{multirow}%
\usepackage{markdown}
\usepackage{amsmath,amssymb,amsfonts}%
\usepackage{amsthm}%
\usepackage{mathrsfs}%
\usepackage[title]{appendix}%
\usepackage{xcolor}%
\usepackage{textcomp}%
\usepackage{manyfoot}%
\usepackage{booktabs}%
\usepackage{algorithm}%
\usepackage{algorithmicx}%
\usepackage{algpseudocode}%
\usepackage{listings}%
\usepackage{xcolor}
\usepackage{pdfpages}

\lstdefinestyle{bashstyle}{
  language=bash,
  basicstyle=\ttfamily\footnotesize,
  keywordstyle=\color{blue},
  commentstyle=\color{gray},
  showstringspaces=false,
  columns=fullflexible,
  keepspaces=true,
}

\usepackage{booktabs,tabularx}

\theoremstyle{thmstyleone}%
\theoremstyle{thmstyletwo}%

\theoremstyle{thmstylethree}%

\begin{document}

\title[Article Title]{Autonomous Code Evolution Meets \textit{NP-Completeness}}


\author*[1,2]{\fnm{Cunxi} \sur{Yu}}\email{cunxiyu@umd.edu}

\author[3]{\fnm{Rongjian} \sur{Liang}}

 \author[4]{\fnm{Chia-Tung} \sur{Ho}}

\author[3]{\fnm{Haoxing} \sur{Ren}}
 

\affil*[1]{\orgdiv{} \orgname{NVIDIA Research}, \orgaddress{\street{}\city{College Park}, \postcode{20740}, \state{MD}, \country{USA}}}

\affil*[2]{\orgdiv{} \orgname{University of Maryland}, \orgaddress{\street{}\city{College Park}, \postcode{20742}, \state{MD}, \country{USA}}}

\affil[3]{\orgdiv{} \orgname{NVIDIA Research}, \orgaddress{\street{}\city{Austin}, \postcode{78717}, \state{TX}, \country{USA}}}

\affil[4]{\orgdiv{} \orgname{NVIDIA Research}, \orgaddress{\street{}\city{Santa Clara}, \postcode{95051}, \state{CA}, \country{USA}}}

\abstract{

Large language models (LLMs) have recently shown strong coding abilities, enabling not only static code generation but also iterative code self-evolving through agentic frameworks. Recently, AlphaEvolve \cite{novikov2025alphaevolve} demonstrated that LLM-based coding agents can autonomously improve algorithms and surpass human experts, with scopes limited to isolated kernels spanning hundreds of lines of code. Inspired by AlphaEvolve, we present SATLUTION, the first framework to extend LLM-based code evolution to the full repository scale, encompassing hundreds of files and tens of thousands of lines of C/C++ code. Targeting Boolean Satisfiability (SAT), the canonical NP-complete problem and a cornerstone of both theory and applications. SATLUTION orchestrates LLM agents to directly evolve solver repositories under strict correctness guarantees and distributed runtime feedback, while simultaneously self-evolving its own evolution policies and rules. Starting from SAT Competition 2024 codebases and benchmark, SATLUTION evolved solvers that decisively outperformed the human-designed winners of the SAT Competition 2025, and also surpassed both 2024 and 2025 champions on the 2024 benchmarks.
}


\pacs[Tribute]{We are deeply grateful to the SAT solving community for nearly three decades of foundational work, which has produced modern SAT solvers capable of handling industrial-scale instances. In particular, the SAT Competition series, founded in 2002, has provided a rigorous benchmarking arena that continues to motivate and accelerate solver innovation, setting high standards for performance and reproducibility in the field.   We further acknowledge the landmark contributions of numerous solvers, such as \emph{zChaff} (Sharad Malik \textit{et al.} \cite{moskewicz2001chaff}), \emph{MiniSat} (Niklas Eén and Niklas Sörensson) \cite{sorensson2009minisat}, \emph{Glucose} (Laurent Simon and Gilles Audemard) \cite{audemard2018glucose}, \emph{Lingeling} and \emph{CaDiCaL} (developed by Armin Biere) \cite{biere2013lingeling, fleury2020cadical}, \emph{MapleSAT} (Vijay Ganesh \textit{et al.}), \emph{Kissat} solver (Armin Biere \textit{et al.}) \cite{biere2024cadical}, \emph{Kissat-MAB} by \citet{cherif2021kissat,zheng2022combining,chen2025kissat_corephase_coreward}, etc.   
The development of \textsc{SATLUTION} was only possible thanks to the accumulated knowledge and the community’s culture of open scientific exchange, exemplified by the open-source availability and the rigorous evaluation practices established by the SAT Competitions. We offer our sincere thanks to the entire SAT solver community for the invaluable foundation they have built. The present work builds directly upon this foundation.
}


\maketitle

\section{Introduction}
The Boolean satisfiability (SAT) problem, first proven NP-complete by \citet{cook2023complexity} in 1971, has served as a cornerstone of computational complexity, linking a wide array of decision problems to a unified theoretical framework \cite{cook2000p}. Beyond theory, SAT solving has emerged as a cornerstone in domains such as hardware verification, software analysis, planning, and cryptography. The evolution of practical SAT solving began in the early 1990s, with the DPLL framework paving the way for more advanced techniques. Breakthroughs around the millennium introduced Conflict-Driven Clause Learning (CDCL), two-watched literals \cite{marques2009conflict}, dynamic restart policies, clause deletion heuristics, bound variable elimination \cite{davis1960computing}, vivification \cite{li2020clause,davis1960computing}, and more, leading to dramatic gains in solver performance \cite{froleyks2021sat, balyo2017sat, biere2009handbook,moskewicz2001chaff}. The SAT Competition, officially held annually since 2002 has been the central platform benchmarking these advances and stimulating community-driven innovation \cite{SATCompetitionWeb}. It regularly hosts a diverse set of tracks (e.g., sequential, parallel, cloud) and attracts new benchmark submissions from planning, industrial verification, and AI communities, acting as both performance contest and repository of hard challenges \cite{Chen2022,Cherif2021,Zheng2023,shi2025dynamicsat,li2023eda}.

Modern SAT solvers have achieved dramatic performance gains through a lineage of ingenious algorithms and heuristics. The DPLL backtracking algorithm, and its evolved form conflict-driven clause learning (CDCL), underpin most state-of-the-art solvers, which are further enhanced by sophisticated branching heuristics, clause database management, and restart strategies. The annual SAT Competitions \cite{SATCompetitionWeb} chronicle these advances, with winning solvers often hand-tuned by experts to exploit specific instance structures. However, designing a champion SAT solver by hand is yielding diminishing returns and continues to demand a high barrier of domain knowledge due to the field’s complex theories and engineering challenges. The manual, intuition-driven development process struggles to navigate the enormous design space of potential solver designs, ranging from hundreds of parameter settings and heuristic variations to novel algorithmic innovations and intricate combinations thereof. This motivates exploring {autonomous solver development}: {could an AI agent automatically evolve a better and correct SAT solver at full repository level, beyond what human engineering achieves?}

Large Language Models (LLMs) have recently demonstrated remarkable capabilities not only in program synthesis \cite{novikov2025alphaevolve,comanici2025gemini,chen2025heurigym,poldrack2023ai,anthropic2024claude}, but also in advancing scientific discovery across diverse domains such as mathematics and the natural sciences. Recent works highlight their ability to perform symbolic regression \cite{grayeli2024symbolic}, accelerate scientific equation discovery \cite{shojaee2024llm}, and even contribute to new mathematical insights \cite{romera2024mathematical}. 
More recently, coding agents that integrate large language models (LLMs) into iterative improvement loops have emerged. Notably, Google’s AlphaEvolve \cite{novikov2025alphaevolve} demonstrated that an autonomous agent can refine algorithms by repeatedly proposing and evaluating code modifications, achieving breakthroughs in automated algorithm design. Sun et al.~\cite{sun2025automatically} specifically 
explores the automatic optimization of SAT solver heuristic modules, focusing 
on algorithmic kernel improvements on the same scale. This line of progress suggests that an LLM-powered agent could also tackle the challenge of building stronger SAT solvers. However, existing efforts like AlphaEvolve have largely focused on self-contained algorithmic kernels rather than the full engineering complexity of real-world solvers. SAT solvers, in contrast, are large, multi-component systems that demand extensive engineering effort, careful integration of diverse heuristics, and rigorous correctness guarantees. Addressing this challenge requires moving beyond one-off code generation toward evolutionary coding agents capable of managing and improving entire repositories under continuous feedback.


\begin{table}[!htb]
\centering
\label{tbl:comparisons}
\caption{Comparison of capabilities and behaviors between AlphaEvolve \cite{novikov2025alphaevolve} and SATLUTION.}
\begin{tabular}{p{0.45\linewidth} p{0.45\linewidth}}
\toprule
\textbf{AlphaEvolve} & \textbf{SATLUTION} \\
\midrule
evolves a single complete code file & evolves entire repository with hundreds of files and Makefile system \\
\hline
evolves up to hundreds of lines of code & evolves tens of thousands of lines across hundreds of files \\
\hline
supports any language & support any language but focused on C/C++ for SAT solving \\
\hline
evaluates for hours in parallel on accelerators & evaluates on distributed CPU clusters with domain-specific formal proof validation \\
\hline
thousands of LLM samples suffice & tens of iterative evolution cycles \\
\hline
benefits from SOTA LLMs & benefits from SOTA LLMs and agent, plus manually designed verifier and evaluator \\
\hline
rich context and feedback in prompts & rich context that integrates static agent rules, self-evolving agent rules, and formal correctness checks \\
\hline
\bottomrule
\end{tabular}
\end{table}

\begin{figure}[!htb]
    \centering
    \includegraphics[width=\linewidth]{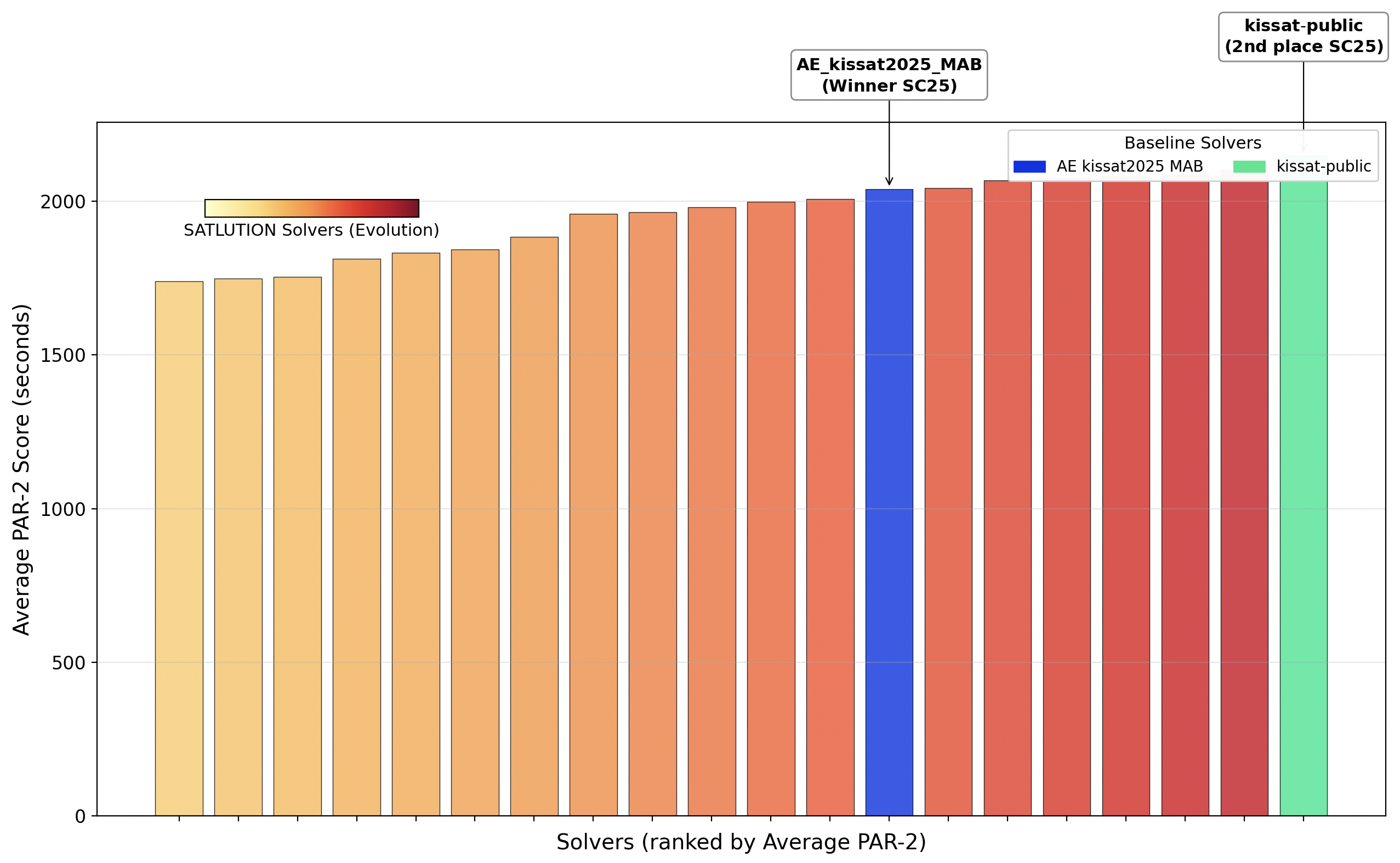}
      \vspace{-3pt} 

    \caption{\textbf{SAT Competition 2025 overall results.} Average PAR-2 score (in seconds) on the SAT 2025 benchmark for SATLUTION and the top solvers from the competition. Lower values are better. SATLUTION attained the lowest PAR-2, outperforming the top-2 2025 winning solvers.}
    \label{fig:2025overall}
\end{figure}

\begin{figure}[!htb]
  \centering
  \includegraphics[width=0.52\linewidth]{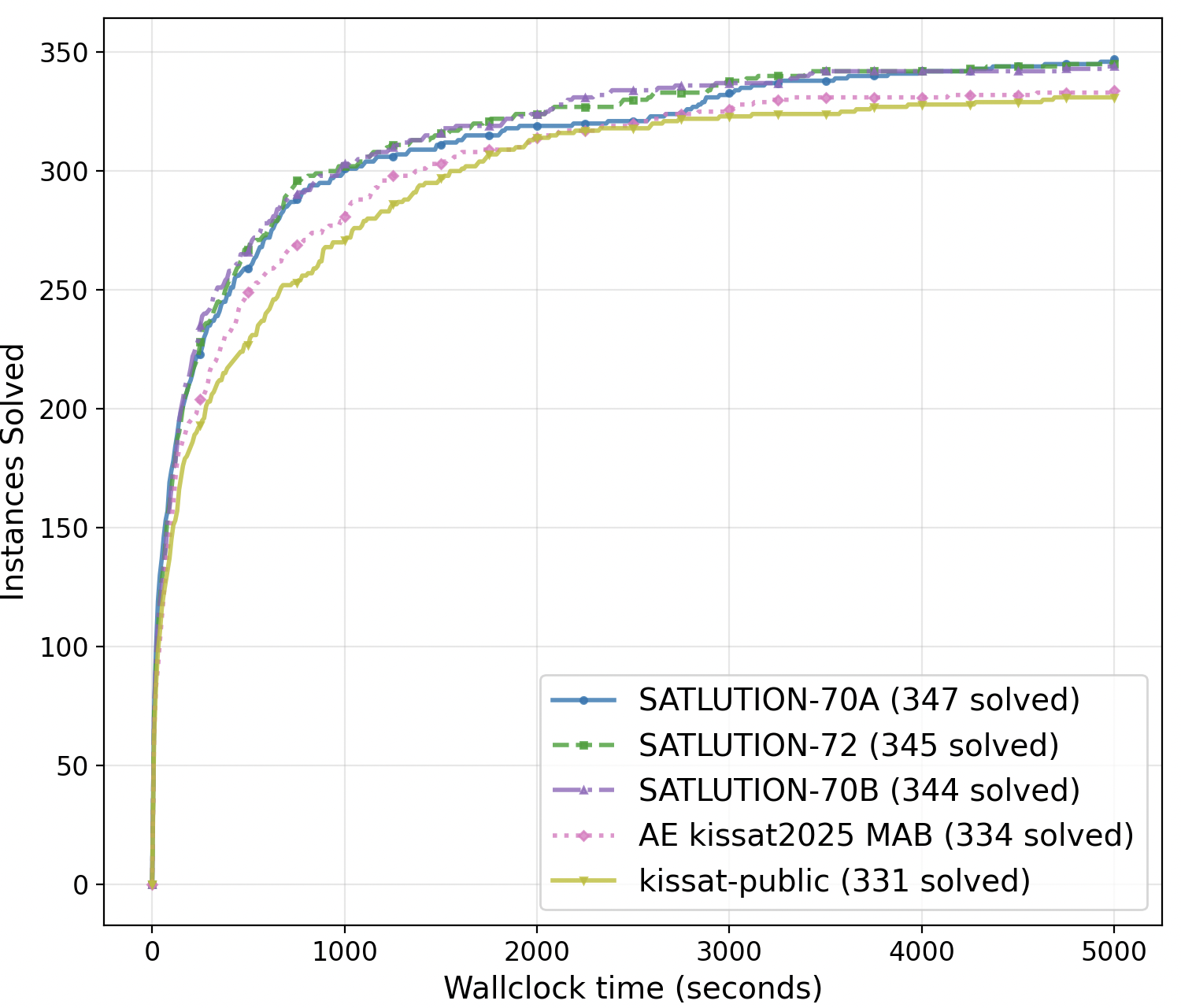} 
  \vspace{-1pt} 
  \caption{Main sequential track comparison plots with SATLUTION best evolved solvers and the winner (\texttt{AE Kissat MAB}) and 2nd place (\texttt{kissat-public}) from SAT Competition 2025 \cite{SATCompetitionWeb}.*Note: The number of solved instances in our evaluation is slightly higher than the official competition release due to differences in evaluator hardware setups. 
    For example, in the official results, \texttt{AE\_KISSAT} solved 327 instances and \texttt{kissat-public} solved 321, whereas in our runs the counts were 334 and 331, respectively, and with averagely faster runtime on all instances. 
    Importantly, despite these absolute differences, the ranking among the top solvers (top-3) remains consistent with the official SAT Competition~2025 results in our test.}
  \label{fig:curves-sc25}
\end{figure}

\begin{figure}[!htb]
    \centering
    \includegraphics[width=1\linewidth]{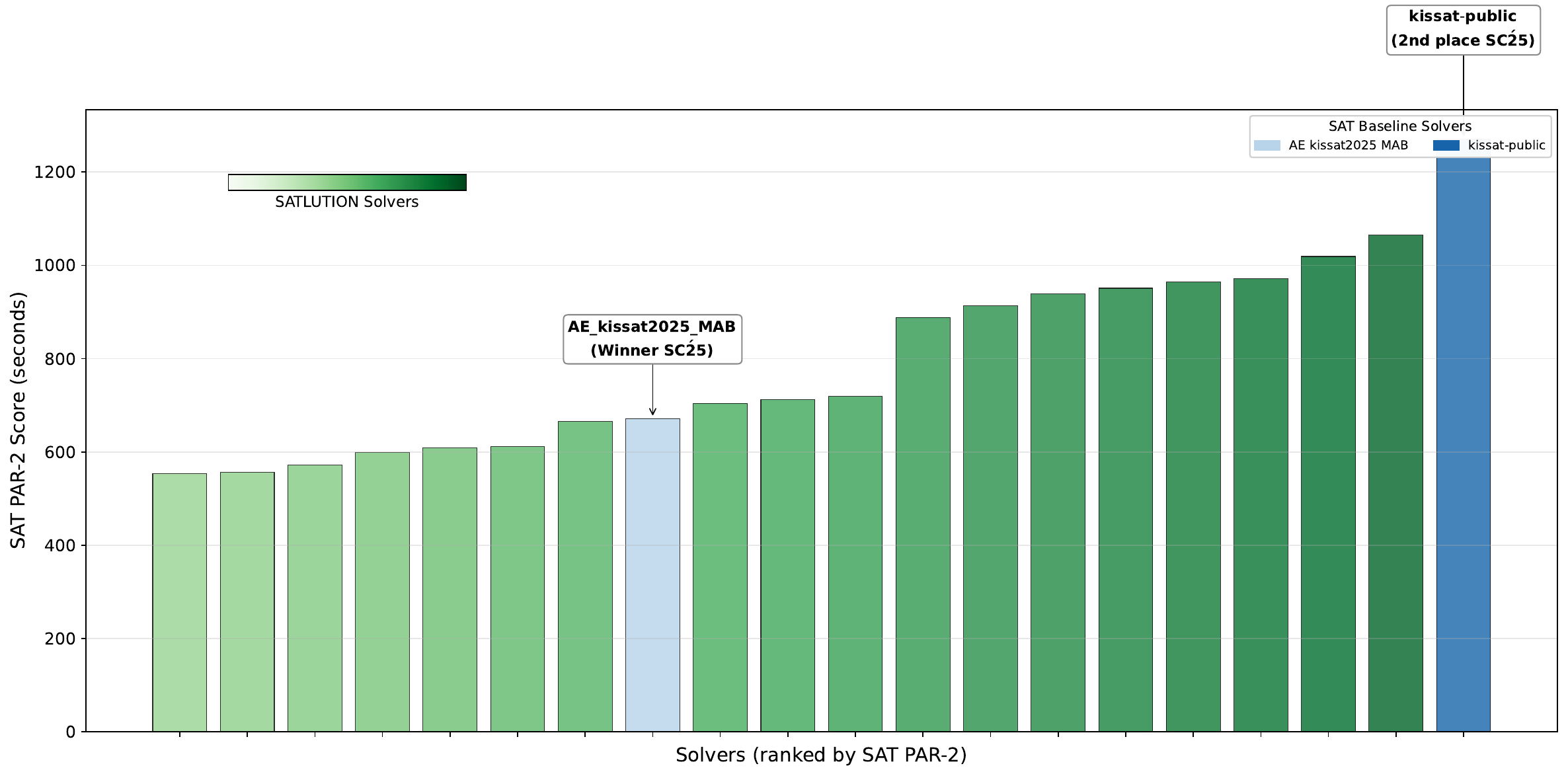}
    \caption{\textbf{SAT Competition 2025 (SAT Category).} PAR-2 performance on \textit{satisfiable} instances. SATLUTION solves SAT instances faster on average than the baseline solvers. 
    }
    \label{fig:2025sat}
\end{figure}

\begin{figure}[!htb]
    \centering
    \includegraphics[width=1\linewidth]{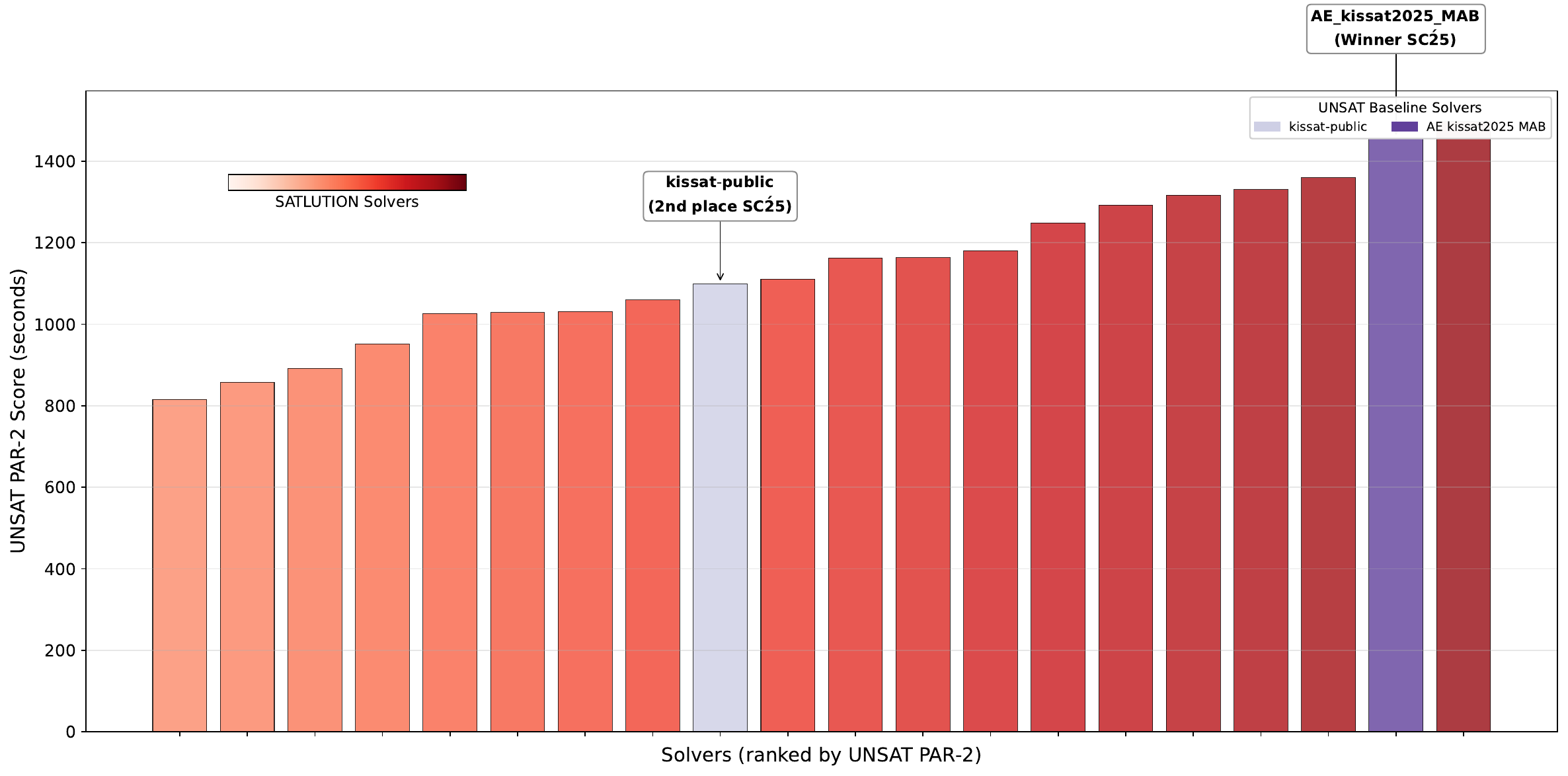}
    \caption{\textbf{SAT Competition 2025 (UNSAT Category).} PAR-2 performance on \textit{unsatisfiable} instances. SATLUTION shows superior efficiency in UNSAT cases as well.}
    \label{fig:2025unsat}
\end{figure}

Here we present \textit{SATLUTION}, a repository-scale, self-evolving coding framework via LLMs focused on SAT solving. SATLUTION orchestrates two LLM-based stages, \textit{Planning} and \textit{Coding}, to iteratively improve a full SAT solver engineering. Starting from a seed portfolio of open-source SAT solvers from the 2024 competition, our agent automatically generates new solver versions, guided by performance feedback on benchmark instances and constrained by rigorous correctness checks. Over dozens of evolution cycles, SATLUTION autonomously discovered solver improvements that \underline{surpass human-designed state-of-the-arts}.  
\textbf{Notably, a family of self-evolved solver outperforms the winning solvers of the SAT Competition 2025 (Figures \ref{fig:curves-sc25}, \ref{fig:2025overall}, \ref{fig:2025sat}, \ref{fig:2025unsat}), using only five selected solvers from SAT 2024 as initial codebase, and with SAT Competition 24 benchmarks as the only feedback.} Our evolution coding framework also shows strong progress in SAT Competition 24 benchmarks (Figures \ref{fig:2024overall}, \ref{fig:2024sat}, \ref{fig:2024unsat}), with a consistent positive progression trajectory (Figure \ref{fig:evolution}). To our knowledge, this work is the first to demonstrate an AI agent based self-evolving coding framework working at repository-level programming, and compete champion-level performance in NP-complete problem solving. SATLUTION is a substantial enhancement over AlphaEvolve (see Table~\ref{tbl:comparisons}), extending LLM-guided agentic evolution from single-file algorithms to full solver repositories. Whereas AlphaEvolve demonstrated iterative refinement of code kernels, SATLUTION leverages LLMs and multi-agent evolution to discover new heuristics, algorithmic innovations, and parameterizations across the entire SAT solver pipeline. This enables not only targeted heuristic improvements but also repository-scale engineering of solvers with correctness guarantees and runtime-driven optimization. 


\section{Results}

\paragraph{SAT Competition 2025 Evaluation.} 
We begin with the strongest demonstration of SATLUTION’s capability: its performance in the SAT Competition 2025. Importantly, all evolutionary training of SATLUTION was conducted exclusively on solver codebases and benchmark instances from SAT Competition 2024 (see Section \ref{sec:methods} \textit{Methods}). Despite never being exposed to the 2025 benchmark set, the evolved solvers achieved new state-of-the-art results in the official 2025 competition. 

Specifically, SATLUTION produced multiple top-ranked solvers that attained the lowest PAR-2\footnote{PAR-2 stands for Penalized Average Runtime with a timeout penalty factor of 2. Lower is better.} scores among all entrants, measured under the competition-standard 5000-second timeout. Figure~\ref{fig:2025overall} summarizes overall performance: SATLUTION solvers achieve substantially lower PAR-2 scores than both the official 2025 winner and all other baselines. In practice, this reflects both a greater number of solved instances and faster runtimes on average.  A breakdown by instance type further highlights this generalization power. On satisfiable instances (Figure~\ref{fig:2025sat}), SATLUTION consistently finds solvers perform the best in the SAT category and similarly on the unsatisfiable instances (Figure~\ref{fig:2025unsat}). Taken together, these 2025 results provide clear evidence that an AI-driven evolutionary process trained on 2024 benchmarks can transfer to and outperform the best human-engineered solvers on the following year’s competition benchmark. This ability to generalize beyond its training environment underscores the central strength of SATLUTION. 

Figure~\ref{fig:curves-sc25} presents a cactus plot comparing the best evolved solvers against the top two official entrants from SAT Competition 2025 (\texttt{AE\_kissat2025\_MAB}, the gold medalist, and \texttt{kissat-public}, the silver medalist). The $x$-axis denotes wallclock time and the $y$-axis the cumulative number of instances solved within that time limit. Across the runtime spectrum, all three evolved solvers consistently solve more instances faster than the competition winners. In particular, SATLUTION developed top-3 solvers solved 347, 345, and 344 instances, compared to 334 and 331 instances for the gold and silver winners, respectively. The separation is most pronounced in the medium-to-hard region ($1000$–$4000$ seconds), where the evolved solvers exhibit stronger scaling and continue to close additional instances while the competition winners plateau. These results confirm that the evolutionary process did not only improve overall PAR-2 scores but also translated into across-the-board runtime advantages, yielding solvers that dominate in both the number of instances solved and the rate at which they are solved.

\begin{figure}[!htb]
    \centering
    \includegraphics[width=\linewidth]{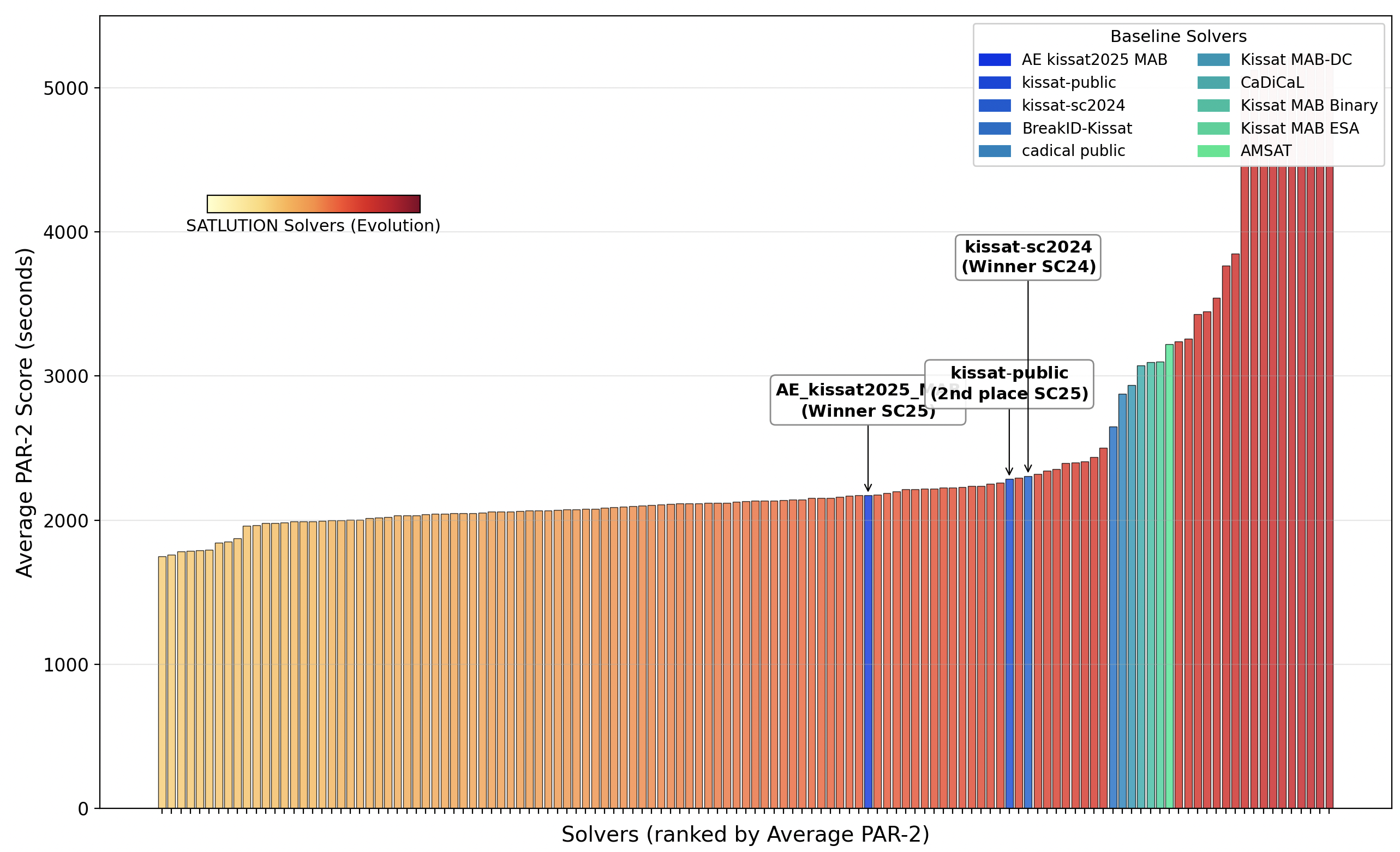}
    \caption{\textbf{Performance on SAT 2024 benchmark.} PAR-2 scores on the SAT Competition 2024 instance set, comparing SATLUTION with baseline solvers from SAT 2024 and 2025. The evolved SATLUTION solver achieves the lowest score, indicating it solves the 2024 instances faster on average than even the original 2024 winner (and the 2025 champion).}
    \label{fig:2024overall}
\end{figure}
\begin{figure}[!htb]
    \centering
    \includegraphics[width=1\linewidth]{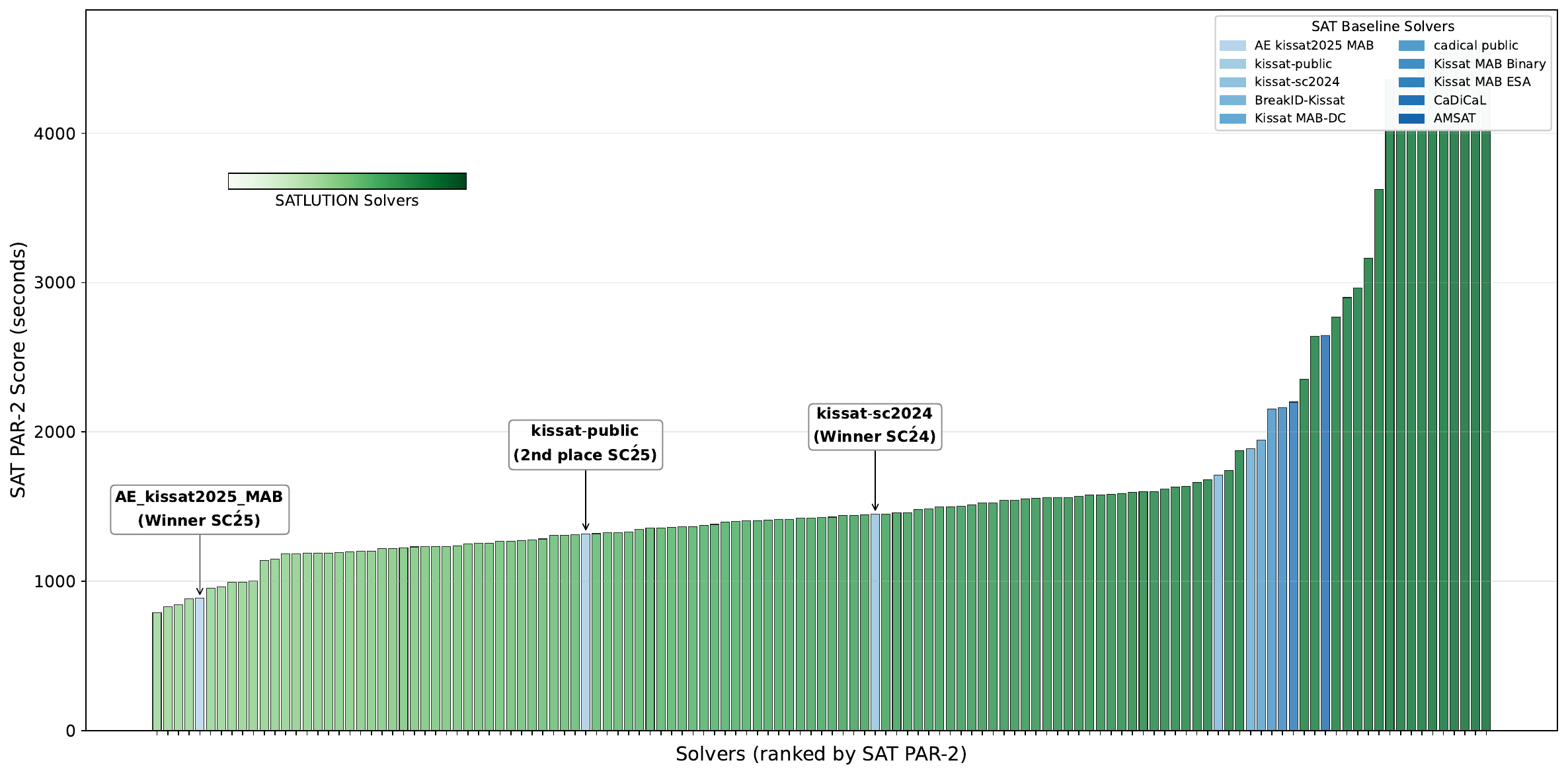}
    \caption{\textbf{SAT 2024 benchmark (SAT-only).} Overall PAR-2 results on satisfiable instances from the 2024 benchmark. SATLUTION shows clear improvement in finding solutions quickly, outperforming the best 2024 solver across these SAT cases.}
    \label{fig:2024sat}
\end{figure}
\begin{figure}[t]
    \centering
    \includegraphics[width=1\linewidth]{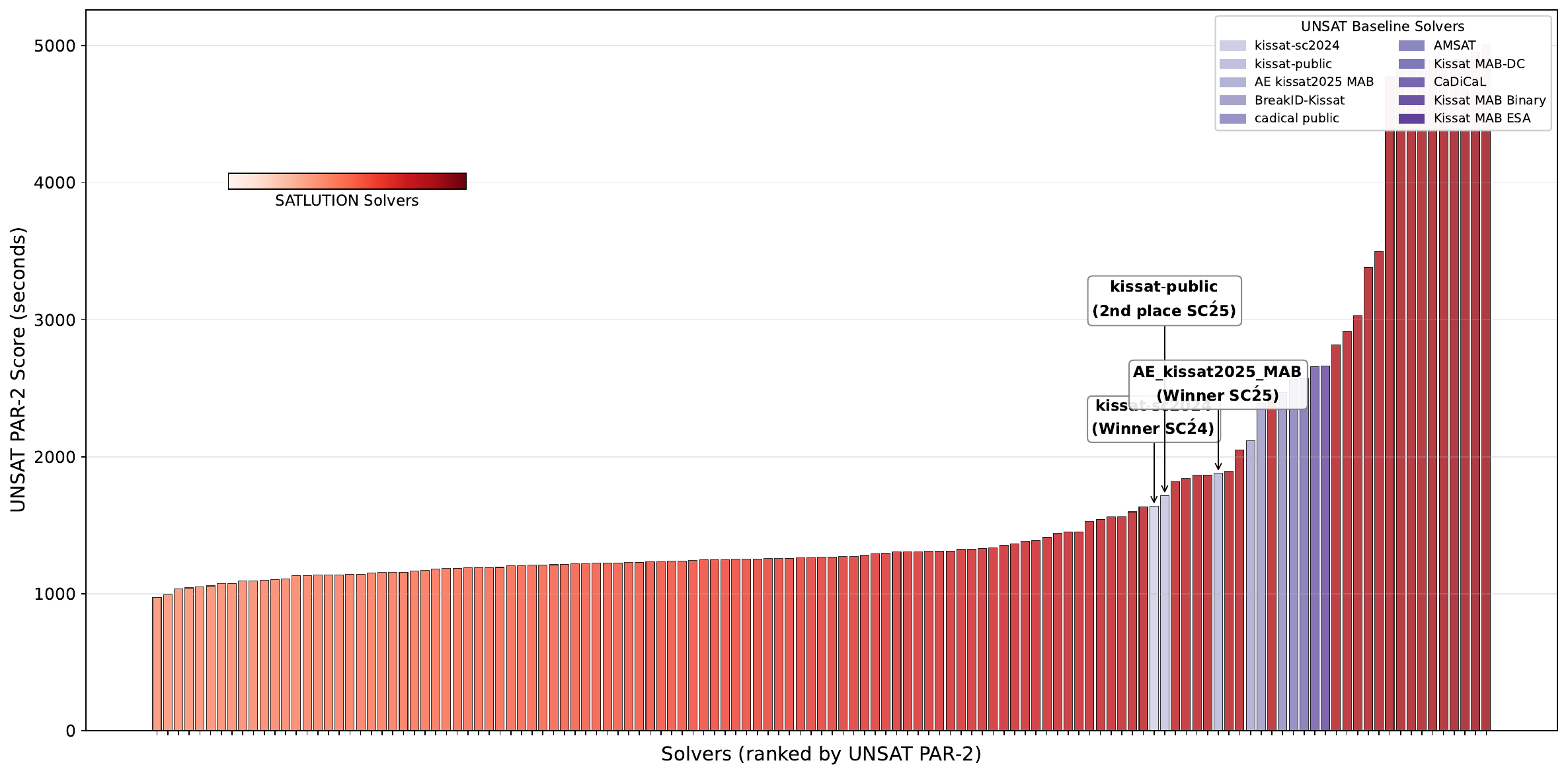}
    \caption{\textbf{SAT 2024 benchmark (UNSAT-only).} PAR-2 results on unsatisfiable instances from the 2024 benchmark. The evolved solver demonstrates superior proof search efficiency, with a lower average time to prove unsatisfiability than the baseline solvers.}
    \label{fig:2024unsat}
\end{figure}

\paragraph{Evolution Results on 2024 Benchmarks} To provide a comprehensive view of the evolution process itself, we next examine SATLUTION’s performance on the SAT Competition 2024 benchmarks, i.e., the very instances used for feedback and our initialized codebase. Over the years, competition-winning solvers have represented the culmination of expert-driven design, with each edition embodying the state-of-the-art at that time. To fairly situate SATLUTION within this progression of human-engineered advances, we therefore include both the official winners of SAT 2024 and SAT 2025 as baselines for comparison.  As shown in Figure~\ref{fig:2024overall}, SATLUTION once again achieves the lowest overall PAR-2 score, substantially outperforming all original 2024 solvers and even surpassing the 2025 champion when run on the 2024 instances. The breakdown by satisfiable and unsatisfiable categories (Figures~\ref{fig:2024sat} and \ref{fig:2024unsat}) further demonstrates that SATLUTION’s improvements are well-balanced across both domains. 

\begin{figure}[!htb]
    \centering
    \includegraphics[width=\linewidth]{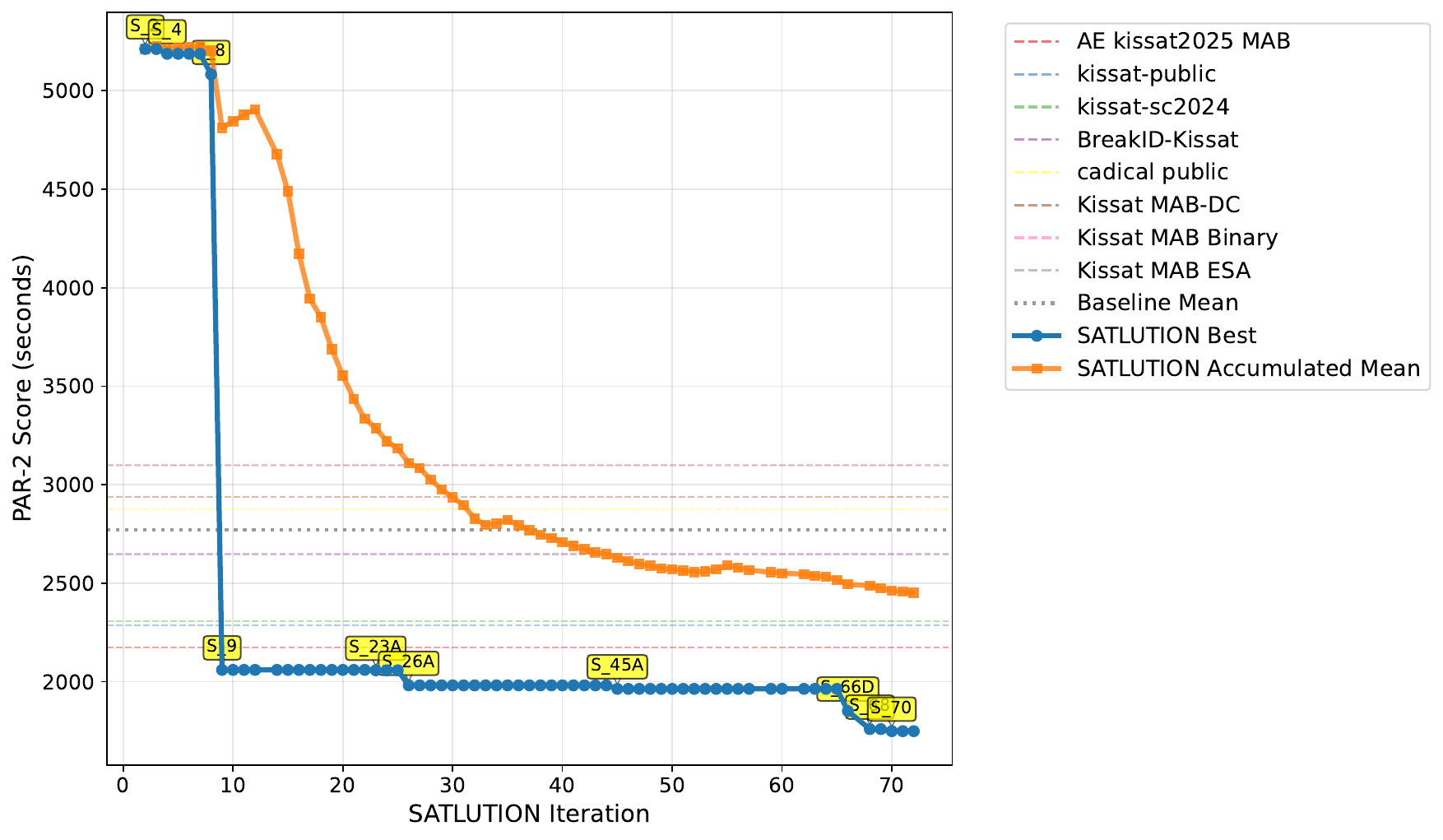}
    \vspace{-3mm}\caption{\textbf{SATLUTION self-evolution performance curve.} PAR-2 score on the SAT 2024 benchmark vs.\ evolution iteration. The curve (blue, solid line) shows the progressive improvement of the solver over 70 iterative cycles. Horizontal dashed lines indicate the PAR-2 performance of top baseline solvers (including the SAT 2024 winner \texttt{kissat-sc2024}, SAT 2025 winner \texttt{AE-kissat-MAB-2025}, etc.) for reference. *Note on failures: across the process, 11 solver variants exhibited partial segmentation faults on subsets of the 400-instance test set, 4 variants failed functional or proof validation, and 9 variants showed significant performance regressions. These failures were automatically detected by the rule–verification pipeline and pruned from subsequent evolution.  }
    \label{fig:evolution}
\end{figure}

\paragraph{Evolution Trajectory} Beyond comparing final solver versions, we analyzed SATLUTION’s entire self-evolution trajectory on the 2024 evolving benchmark. Figure~\ref{fig:evolution} plots the solver’s PAR-2 performance as a function of evolution iteration (cycle), alongside reference lines for the baseline solvers’ performance. The starting point at iteration 0 corresponds to the ensemble of seed solvers (with an initial average PAR-2 around the baseline mean). We observe that SATLUTION made rapid progress in the first 5–10 iterations, quickly lowering the PAR-2 by incorporating obvious improvements (for example, combining complementary strengths from the different seed solvers). Improvements continued in subsequent iterations, albeit with diminishing returns, as the agent tackled more subtle optimizations. By roughly iteration 50, SATLUTION’s solver began outperforming the 2025 human-designed winner on the 2024 benchmark, and by the final iteration (~70) it solidly surpassed all baselines. The lowest achieved PAR-2 is marked as ``SATLUTION Best`` in Figure~\ref{fig:evolution}, which lies well below the prior state-of-the-art. This evolutionary curve illustrates the benefit of automated iterative refinement: the agent was able to steadily accumulate small performance gains into a significant overall advantage. It also highlights the stability of the process, no catastrophic regressions occurred once the correctness safeguards (discussed below) were in place, as evidenced by the monotonic trend of the SATLUTION accumulated mean performance. In summary, the self-driven evolution not only found a final solver that is superior to any single human-designed solver, but it did so via a smooth and reproducible optimization trajectory.

\section{Methods}\label{sec:methods}

\begin{figure}[!htb]
    \centering
    \includegraphics[width=\textwidth]{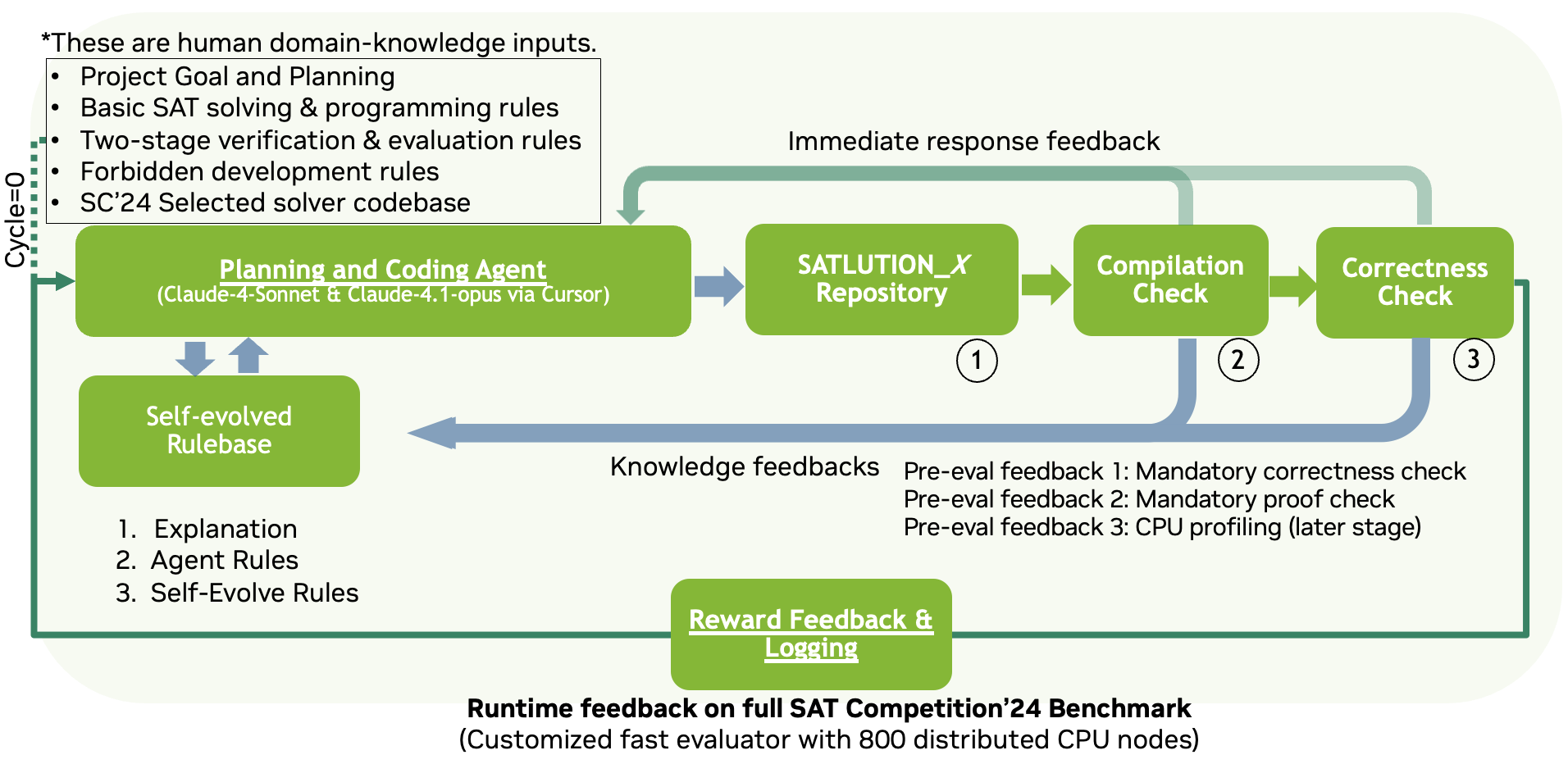}
    \caption{\textbf{SATLUTION self-evolving agent framework.} A schematic of the iterative loop used by SATLUTION. The Planning Agent (left) formulates improvement plans, which the Coding Agent (right) realizes by modifying the solver’s code in the central repository. Each cycle is governed by a rulebase encoding domain knowledge and human-provided constraints (top-left), ensuring the agent’s proposals remain sensible. After code modification, the new solver is subjected to an evaluation on benchmarks using a distributed testbed (far right), and multi-faceted feedback (including correctness checks and performance metrics) is returned to the agents. This feedback loop allows SATLUTION to improve the solver over successive iterations autonomously.}
    \vspace{-3mm}
    \label{fig:framework}
\end{figure}

\paragraph{SATLUTION Architecture} 
We designed SATLUTION as an autonomous agent-based code evolution system that iteratively improves SAT solvers at the repository level (Figure~\ref{fig:framework}). The architecture leverages state-of-the-art LLM agents that integrate both \textit{Planning} and \textit{Coding} capabilities. In the \textit{Planning} stage, the agent performs high-level reasoning: at the initial cycle, it analyzes the SAT Competition 2024 solver codebases and their performance, proposing promising directions for modification such as optimizing specific heuristics or refactoring key modules. In subsequent cycles, it reasons about the accumulated code changes, performance metrics, and observed failures, and then formulates an evolution plan for the next iteration. The \textit{Coding} stage executes these plans by directly editing the solver repository, leveraging the generative coding abilities of LLMs to implement changes at scale. The Coding Agent also manages auxiliary tasks, including updating build system configurations (e.g., Makefiles), fixing compilation errors, and debugging both functional failures (e.g., correctness violations) and execution errors (e.g., segmentation faults).

This architecture is designed to overcome two challenges in solver development we discussed earlier. First, the steep {engineering and domain barriers} are mitigated by embedding simple domain knowledge into initialization rules and constraints, enabling agents to evolve solvers without requiring deep SAT-specific expertise. Second, the {vast design space} of heuristics, parameters, and algorithmic variants is navigated through iterative exploration guided by distributed runtime evaluation, allowing the framework to orchestrate and tune existing strategies while also proposing novel innovations. By coupling LLM coding capabilities with rule-based guardrails and scalable feedback, SATLUTION transforms solver development from a manual, expert-driven process into an autonomous cycle of repository-scale evolution.

\paragraph{Agent Framework Leverages} 
Our agent framework builds on the Cursor environment and two Claude models to support solver self-evolution. A key element is the rule system, which provides structured guidance to the agents across different phases of the workflow. In practice, rules are organized to cover planning, basic C/C++ programming, pre-evaluation verification, post-evaluation reward analysis, and long-term knowledge learning. Planning rules help the planning agent decompose solver modifications and define iteration strategies. Programming rules enforce C/C++ correctness and style, ensuring memory safety and adherence to solver interfaces. Pre-evaluation rules verify code validity before execution, while post-evaluation rules guide the analysis of runtime feedback, distinguishing performance gains from regressions. Knowledge-learning rules maintain institutional memory, including forbidden rules that block known bad patterns, champion rules that preserve strategies from top-performing solvers, and failure rules that prevent repetition of past mistakes. Cursor’s rule system enables fine-grained scoping, with rules that can be always applied, automatically attached, or manually triggered depending on the phase of evolution. Within this setup, the planning Claude model generates modification blueprints, while the coding Claude model, operating under programming and verification rules, implements changes in C/C++. We also observed that agents without initial static rule guidance, or those relying solely on dynamically self-evolved rules, consistently underperformed: lacking a stable baseline, they produced noisy or unstable modifications that failed to sustain evolutionary progress. The combination of initial rule-guided scaffolding and subsequent self-evolved refinements proved essential for effective solver evolution, ensuring that each cycle was both disciplined and adaptive.

\begin{table}[!htb]
\centering
\label{tbl:staticrule}
\caption{Core components described in the static rulebase (in markdown) of the SATLUTION evolution framework. The components were originally written by the authors and subsequently refined through prompt tuning before being used as the static initialization rule for guiding the agents.}
\begin{tabular}{p{0.21\linewidth} p{0.70\linewidth}}
\toprule
\textbf{Component} & \textbf{Description} \\
\midrule
\textbf{0. Project Goal} & Description of the goal of repository-scale self-evolution of SAT solvers using LLM agents, targeting state-of-the-art performance while ensuring correctness and robustness. \\
\hline
\textbf{1. Domain Knowledge Initialization} & Embeds basic SAT heuristics such as restart policies, preprocessing, in-process simplification, and core CDCL algorithmic principles as static rules, as well as the soundness checks must remain intact, proof validation required. Provides a foundation for agents to evolve solvers without deep SAT-specific expertise. \\
\hline\hline
\textbf{2. Critical Correctness Rules} & Enforces strict guarantees: no instance-specific optimizations; definitive SAT/UNSAT outputs only; mandatory DRAT proof generation for UNSAT; zero tolerance for predictive or heuristic-only termination. \\
\hline\hline
\textbf{3. Repository Structure \& Tracking} & Each solver variant (\texttt{SATLUTION\_x}) must include: \texttt{HYPOTHESIS.md} (agent’s proposed modifications and rationale), \texttt{CHANGELOG.md} (implemented changes), and \texttt{RESULTS.md} (evaluation conclusions). Ensures complete lineage, traceability, and reasoning transparency. \\
\hline\hline
\textbf{4. Pre-Evaluation Verification} & Mandatory correctness gate before full evaluation: compilation check, run on 115 CNF tests, SAT/UNSAT output verification, proof validation, and multi-baseline agreement (e.g., vs.\ kissat-sc2024). Any mismatch halts evolution immediately with feedbacks. \\
\hline\hline
\textbf{5. Evaluation Protocol} & After verification, solvers are benchmarked using manually developed distributed evaluator \texttt{automated\_evaluator.py} on SAT Competition datasets. Metrics include solved instances, runtime, memory, and PAR-2 (overall, SAT-only, UNSAT-only). Results are analyzed via \texttt{post\_evaluation\_analyzer.py} with correctness integration that provides rich context to trigger agent reasoning. \\
\hline\hline
\textbf{6. Self-Evolving Methodology} & Iterative \textit{Generate $\rightarrow$ Compile $\rightarrow$ Test $\rightarrow$ Analyze \& Reason $\rightarrow$ Evolve} loop under a Champion/Challenger model. Agent proposes directions, todo tasks of the five steps, and propose and make code edits; distributed runtime feedback guides evolution. Successful challengers replace champions, creating a continuous self-improvement cycle. \\
\bottomrule
\end{tabular}
\end{table}

\begin{lstlisting}[style=bashstyle,
                   caption={Repository structure mandantory rule for SATLUTION evolution process.},
                   label={lst:satlution-structure}]
SATLUTION_x/                  # Root folder of x cycle evolution
-- bin/                       # Executable directory
---- solver_binary            # Main solver executable
---- starexec_run_default     # Execution flow in Bash script
-- src/                       # Complete source code
-- build/                     # Build scripts and Makefile
-- CHANGELOG.md               # Modifications from previous version
-- HYPOTHESIS.md              # Performance expectations and rationale
-- RESULTS.md                 # Benchmarking outcomes
-- starexec_build             # Root build script; run: bash starexec_build
\end{lstlisting}

\paragraph{Initialization Setup} At the initial iteration (cycle 0), the repository was seeded with five high-performance SAT solvers from the SAT Competition 2024: \texttt{kissat-sc2024}, \texttt{kissat-mab-dc}, \texttt{kissat-mab-binary}, \texttt{BreakID-Kissat}, and \texttt{AMSAT}\footnote{Portfolio or parallel 
solver construction was explicitly forbidden in the SATLUTION evolution. 
All evolved solvers adhere strictly to the sequential solver requirements 
of the SAT Competition.}. These solvers provided a diverse starting pool of algorithms and heuristics, including alternative branching strategies, clause management techniques, and, in the case of BreakID (a symmetry-breaking preprocessor). Instead of selecting a single baseline, we exposed SATLUTION to all five codebases, enabling it to hybridize strategies and draw from complementary strengths. These solvers were selected based 
on the authors’ limited internal knowledge of solver implementations and the 
official 2024 competition results.

Alongside the codebases, we provided the system with the initial evolution rulebase (Table~\ref{tbl:staticrule}). This rulebase encodes our domain knowledge and establishes the critical correctness requirements (e.g., soundness of SAT/UNSAT results, proof validation), mandatory repository structure, and documentation rules. Each solver variant produced during evolution must strictly follow the enforced structure shown in Listing~\ref{lst:satlution-structure}, which serves as the first validation step (\textcircled{1} in Figure~\ref{fig:framework}). Every iteration is stored in its own directory (\texttt{SATLUTION\_x/}), with a standardized layout: a \texttt{src/} folder for source files, a \texttt{bin/} folder containing the compiled binary and a competition-compliant \texttt{starexec\_run\_default} script, and a top-level \texttt{starexec\_build} script for reproducible compilation. Documentation is enforced at each cycle via \texttt{HYPOTHESIS.md} (planned modifications and performance rationale), \texttt{CHANGELOG.md} (actual code edits), and \texttt{RESULTS.md} (evaluation outcomes and reasoning).  Finally, we imposed additional programming constraints to keep the task focused on algorithmic innovation, such as solvers had to remain single-threaded C/C++ programs, disallowing external libraries or non-standard APIs. Together, this initialization setup provided SATLUTION with both a diverse solver foundation and a structured, rule-governed environment for autonomous repository-scale evolution.

\begin{figure}[t]
    \centering
    \includegraphics[width=1\textwidth]{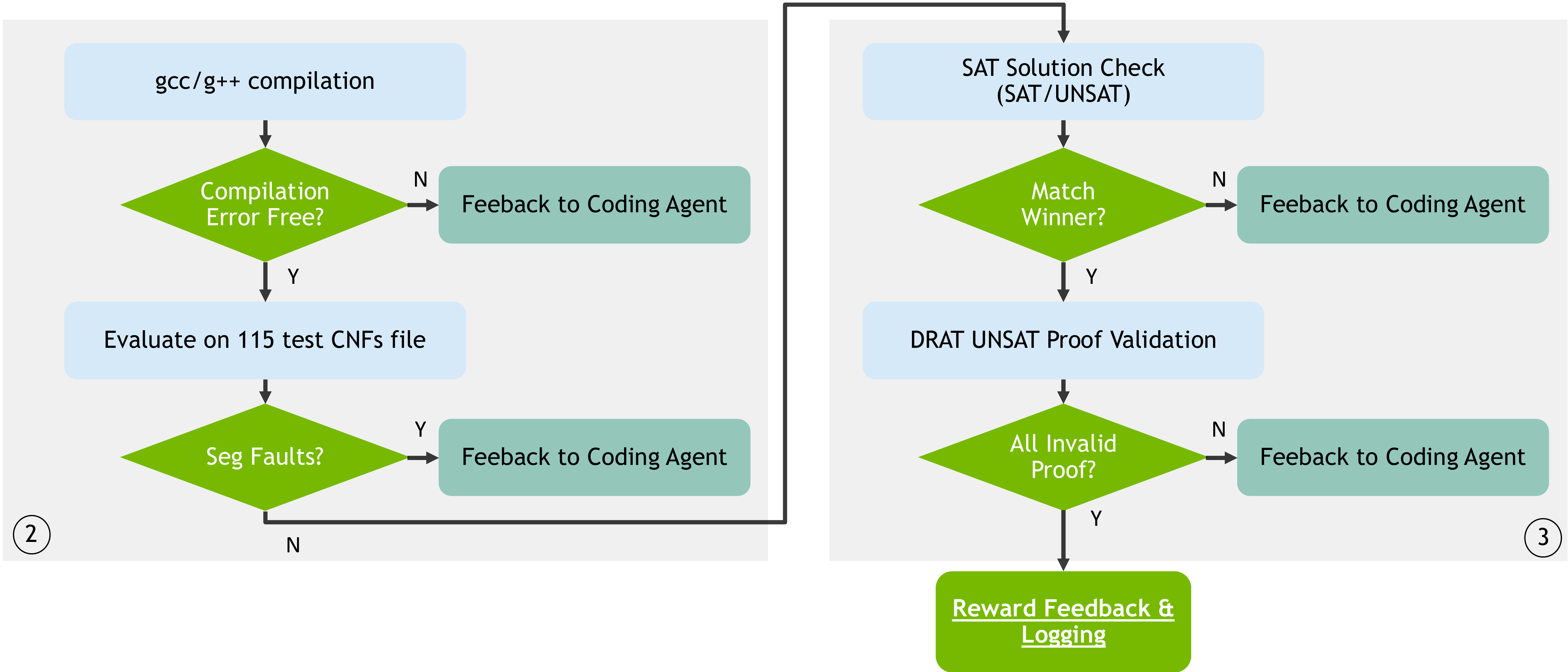}
    \caption{Detailed illustration of the \textbf{two-stage verification pipeline for solver correctness (\textcircled{2} and \textcircled{3} in Figure \ref{fig:framework}).} After each code modification, Stage 1 performs a compilation check and runs a small set of smoke-test instances to catch crashes or blatant errors (e.g., memory faults or incorrect SAT/UNSAT responses on trivial cases). If Stage 1 passes, Stage 2 rigorously verifies the solver’s results on a broader set of instances: any reported SAT result must come with a valid satisfying assignment (checked against the formula), and any UNSAT claim must be accompanied by a valid DRAT proof (validated by an external checker). Failures at either stage trigger feedback to the Coding Agent, which then attempts to repair the solver in the next iteration. Only a solver that clears both stages is considered correct and eligible for performance evaluation. *Stage 1 verification feedback is crafted by the agent automatically, and Stage 2 feedback flow is manually crafted by the authors instead of agents.}
    \label{fig:verification}
\end{figure}

\paragraph{Verification and Correctness Safeguards} Evolving a complex solver automatically risks introducing bugs or unsound behavior. To prevent this, each iteration of SATLUTION undergoes a stringent two-stage verification pipeline (Figure~\ref{fig:verification}) before its performance feedback is assessed. \textbf{Stage 1} (\textcircled{2} in Figure \ref{fig:framework}) is a basic \textit{compilation and smoke test}: after the Coding Agent produces a new code version, the system attempts to compile the solver with standard settings. Any compile error immediately flags the iteration as failed, prompting feedback to the Coding Agent to fix the issue in the next attempt. If compilation succeeds, the solver is run on a small set of 115 trivial CNF formulas from \texttt{kissat} repository, designed to exercise basic functionality (e.g., unit propagation, simple satisfiable and unsatisfiable cases). We monitor for runtime errors like crashes (segmentation faults, assertion failures) or obviously incorrect answers on this smoke test. If any such issue is detected, the iteration is halted and the agent is informed of the failure (with details, such as ``segmentation fault in preprocessing`` or ``incorrect result on a known satisfiable instance``), so it can attempt a corrective change. Only if the new solver passes {Stage 1} (compiles and shows no immediate errors) do we proceed to \textbf{Stage 2} (\textcircled{3} in Figure \ref{fig:framework}), the full correctness validation. In Stage 2, the solver is tested on a larger set of benchmark instances with known outcomes (drawn from past competition problems). For each instance solved, we perform an exacting check: if the solver reports ``\texttt{SAT}``, we verify the returned assignment satisfies the formula; if it reports ``\texttt{UNSAT}``, we verify the produced proof using an external DRAT proof checker. Any deviation (such as a wrong answer or an invalid proof) triggers a failure, and the agent is penalized in its feedback for producing an incorrect solver. Only if all Stage 2 checks are passed is the solver deemed correct and allowed to receive a positive reward based on its performance. This two-tiered approach (fast shallow tests followed by deep validation) ensured that SATLUTION’s evolution never veered into unsound territory — throughout our experiments, no solver version that passed both stages ever produced a misclassification on the competition benchmarks. The strict correctness gatekeeping inevitably slows down the evolution process (since many candidate changes are rejected), but it is essential for safety given the high stakes of automated code modifications in a theorem-proving context.

\paragraph{Performance Evaluation and Feedback Metrics} Once a new solver version is verified as correct, SATLUTION evaluates its performance to guide the next cycle of improvements. We deployed a \textbf{distributed runtime evaluator} harness to obtain detailed performance feedback efficiently. In each iteration, the solver is run on the full SAT Competition 2024 benchmark suite (400 instances across application and hard combinatorial categories) using a cluster of 800 CPU nodes in parallel. This massive parallelism allows us to gather results from an entire benchmark run in roughly an hour, providing the agent with near-immediate feedback on how the latest changes affected performance. The feedback includes a variety of metrics (Table~\ref{tab:agent-metrics}) designed to give a comprehensive view of solver behavior. These include: the number of SAT and UNSAT instances solved (and whether those match the Virtual Best Solver’s known outcomes for each instance), breakdowns of how many instances are solved within certain time cutoffs (300s, 600s, 1000s, etc., which helps identify if changes favor quick wins versus long-haul performance), and memory usage statistics. The Virtual Best Solver (VBS) is essentially an oracle that for each instance knows the correct answer and the best result achieved by any baseline solver, allowing us to flag any misclassification or uniquely solved instances. The primary metric driving evolution is the PAR-2 score — the competition-standard penalized average runtime, which heavily penalizes unsolved instances by assigning a 5000-second cost for each timeout. SATLUTION’s reward in each cycle is largely based on the inverse of PAR-2 (lower PAR-2 = higher reward), encouraging the agent to both solve more instances and reduce runtimes. However, we found it beneficial to supplement this single score with the additional metrics in Table~\ref{tab:agent-metrics} for richer feedback. For example, if a code change significantly improved SAT solving speed but caused a regression on UNSAT cases, the composite feedback would reflect that nuance (the agent might see an increase in “\# solved SAT instances” but a drop in “match with VBS” if it started mislabeling some UNSAT as SAT due to incomplete search, etc.). By exposing multiple facets of performance, the Planning Agent can make more informed decisions about which trade-offs are acceptable. In practice, we observed the agent learning to balance these factors, yielding solver improvements that were broad (solving more of both SAT and UNSAT problems) rather than narrowly optimizing one category at the expense of the other.

Beyond headline competition results, our experiments also reveal important insights into how feedback design shapes solver evolution. In the early iterations, when SATLUTION was guided primarily by the metric of ``total number of instances solved,'' the evolved solvers tended to focus on strengthening UNSAT performance. This is consistent with the fact that unsatisfiable instances are often bottlenecks, and solving or proving more of them directly increases the total solved count. However, once we incorporated PAR-2 scores together with finer-grained metrics such as solved-instance distributions across multiple time cutoffs (see Table~\ref{tab:agent-metrics} in Section~\ref{sec:methods}), SATLUTION shifted its emphasis towards satisfiable instances. In particular, the agent began to optimize the speed of solving medium-to-hard SAT instances, where improvements contribute disproportionately to reducing the overall PAR-2 score.  A manual analysis of PAR-2 dynamics further clarifies this effect: SAT performance carries slightly greater weight than UNSAT performance in the aggregate score. In other words, if a solver achieves significant gains on SAT PAR-2 while losing slightly on UNSAT PAR-2, the overall PAR-2 score still improves. This phenomenon was also reflected in the official SAT Competition 2025 results: the evolved solver \texttt{AE\_kissat\_MAB} exhibited superior performance on satisfiable instances compared to \texttt{kissat-public}, while trailing slightly on unsatisfiable cases, yet the overall PAR-2 outcome secured its first-place victory. These observations highlight the importance of feedback signal design in guiding solver evolution, and suggest that different reward formulations may be tuned to emphasize specific strengths depending on application needs.

\begin{table}[t]
    \centering
    \caption{Agent feedback metrics for SAT-solving performance. VBS = Virtual Best Solver across all compared solvers (provides ground-truth outcomes for instances). These metrics are computed after each iteration’s solver is run on the full benchmark and are used to guide the agent’s next steps. *The evaluator and feedback/rewards processor are manually crafted by the authors instead of agents.}
    \label{tab:agent-metrics}
    \begin{tabular}{|p{3cm}|p{7cm}|}
    \hline
    \textbf{Metric} & \textbf{Description} \\
    \hline
    Match with VBS (SAT/UNSAT) & Whether the solver’s answers (SAT or UNSAT) match the known correct outcome for every instance. Any discrepancy indicates an incorrect result (trigger failure learning rule generation). \\
    \hline
    \# solved SAT instances & Number of instances correctly solved (satisfiable cases) within the time limit. \\
    \hline
    \# solved UNSAT instances & Number of instances correctly solved (unsatisfiable cases proven) within the time limit. \\
    \hline
    \# solved $\leq 300$s, $\leq$600s, $\leq$1000s, $\leq$2000s, $\leq$3000s, $\leq$4500 & Count of instances solved within 5 specific time thresholds, providing a runtime distribution of solved instances (lower thresholds indicate how many easy instances are quickly solved, higher thresholds reflect ability on hard instances up to the 5000s cutoff). \\
    \hline
    Average memory usage & Mean peak memory usage of the solver across all instances, in gigabytes. \\
    \hline
    Max memory usage & Peak memory usage observed on any single instance, in gigabytes. This helps catch any memory blow-ups or leaks introduced by changes. \\
    \hline
    Additionally solved vs. baselines & How many instances the new solver solved that none of the baseline solvers could solve. This highlights novel strengths introduced by the agent (solving previously unsolved instances). \\
    \hline
    Overall, SAT, and UNSAT PAR-2 scores  & PAR-2 score feedback. Note that PAR-2 score feedback is used only in the later evolution cycles (after 33 cycle) to avoid the agent to ``trick`` and generate ``wired`` optimizations with a clear competition scores. The intuition is to let SATLUTION purely improving overall SAT solving performance instead of winning in PAR-2 score. \\
    \hline
    \end{tabular}
\end{table}

\paragraph{Rule System: Initialization and Self-Evolution}

 A central takeaway from our study is that the rule system significantly improves the efficiency of solver evolution. Intuitively, the static and dynamic rulebase provides enforced planning guidance and prunes the search space, reducing wasted exploration on development or debugging failures and steering the agent toward meaningful algorithmic innovations.  

\vspace{3mm}
\noindent
\textit{Initialization rules that encode human knowledge} -- 
Given limited domain-knowledge that the authors have to alter core SAT internals without risking soundness, SATLUTION begins with an explicit, static rulebase that formalizes {what a correct SAT solver must do} and {how proposals will be verified and evaluated}. Concretely, as shown in Table \ref{tbl:staticrule}, (1) {critical correctness constraints}: any change applied to the solver repository must {not} bypass the mandatory solution/proof checks; (2) {verifier binding}: every candidate build {must} pass a two-stage pipeline -- compile/segfault gate; SAT/UNSAT validation (SAT model check; UNSAT DRAT proof), and only then enter performance evaluation; (3) {evaluator binding}: the agent is told precisely how all metrics (see Table \ref{tab:agent-metrics}) will be computed and how SAT vs.\ UNSAT subsets are scored and reported; (4) {safe-coding and process rules}: memory safety (no ad-hoc static buffers or non-allocator misuse), modularity, mandatory logging (\texttt{CHANGELOG.md}, \texttt{HYPOTHESIS.md}, \texttt{RESULTS.md}); (5) {seed knowledge}: the prompt includes a summary of SAT Competition 2024 standings for the five seed solvers (kissat-sc2024, kissat-mab-dc, kissat-mab-binary, breakid-kissat, amsat), including their PAR-2 ranks and instance counts in SAT/UNSAT, so the agent has a grounded starting point for trade-off considerations and can reference strong baselines when proposing edits. 

 \vspace{3mm}
\noindent
\textit{With vs.\ without static rules: why rules matter} -- 
In ablations where we removed the static initialization rules, the agent explored {plausible but misleading} directions, e.g., early termination heuristics that spuriously predict satisfiability, or edits that accidentally weaken proof obligations—leading to frequent DRAT/solution check failures, degenerate optimizations on easy SAT instances, or non-reproducible speed-ups. For example, in our preliminary exploration stage, with 20 cycles of evolution without the static rulebase, there are 7 failure cases, and 8 cycles produce solvers that significantly degraded performance. With the static rules in place, the system sort of prunes the evolution directions and space, e.g., forbidden patterns are flagged, failing builds are discarded before evaluation, and documentation/logging discipline reveals regressions quickly. 

\vspace{3mm}
\noindent
\textit{Dynamic self-evolved rules in the loop} -- Based on the above study, we realize that the rulebase to the agent flow is critical
to ensure the efficient self-evolution process. Therefore, we propose to let the rulebase self-evolve as well, such that the rules themselves keep pace with what the agent learns. Specifically, we embed rulebase self-evolving mechanisms in every iteration: {rule\_compliance\_verification} and {automatic\_rule\_evolution}.  
{(1) Rule compliance verification:}
Before the agent’s proposal enters the verifier/evaluator, a compliance script suite enforces: presence of mandatory docs (\texttt{CHANGELOG.md}, \texttt{HYPOTHESIS.md}, \texttt{RESULTS.md}), packaging/branding correctness (e.g., \texttt{bin/starexec\_run\_default}), and {forbidden pattern} scans across \texttt{.c/.h} (e.g., unsafe static state, non-standard memory functions outside vetted allocator paths). Recommended pre-commit hooks run these checks automatically; weekly audits ensure drift is detected, and cursor is periodically queried to confirm it {cites and enforces} rule numbers and files (e.g., “Rule 01 pre-evaluation testing,” “Rule 03 mandatory logging,” “Rule 04 forbidden patterns”). This converts rules from passive text into active gates with measurable effectiveness (prevention rate, compliance score, time-to-fix, and coverage).  {(2) Automatic rule evolution:}  
After each cycle, a post-mortem analyzer parses compile/link errors, verifier mismatches, missing-documentation events, and {new} failure signatures (e.g., predictive termination, heuristic result short-cuts), then proposes rule patches. An update engine injects those as concrete diffs, e.g., {append a new ``FORBIDDEN`` block to Rule 04 with a code snippet and rationale}, tighten Rule 01 pre-evaluation tests, strengthen Rule 03 logging, and record a versioned snapshot of the rule set. 
The initialization rule set explicitly seeded two-step requirements, which in turn {generated and maintained} a concrete family of rule files and checkers in \texttt{.cursor/rules/}:
\begin{quote}\small
\texttt{00\_rule\_compliance\_verification.md}, \texttt{01\_pre\_evaluation\_testing.md}, \texttt{02\_critical\_correctness.md}, \texttt{03\_mandatory\_logging.md}, \texttt{04\_forbidden\_patterns.md}, \texttt{05\_automatic\_rule\_evolution.md}.
\end{quote}
These files evolve with cycles and remain machine-actionable via scripts (compliance checker, post-cycle analyzer, rule updater, version manager). The result is a rule system that {co-evolves} with the solver evolution: for example, each new failure pattern is rapidly codified as prevention; each process gap hardens pre-evaluation or documentation; and cursor’s guidance reflects the newest rule version by design.   
Operationally, only rule-compliant proposals reach the mandatory Stage-2/Stage-3 checks (compile/segfault gate; SAT model validation; UNSAT DRAT proof) and, if successful, proceed to distributed runtime evaluation for PAR-2 and per-category (SAT/UNSAT) analysis. 

\paragraph{Knowledge Components Learned through Evolution}

\begin{table}[!htb]
\centering
\caption{Distilled solver-design insights and techniques, ordered by development flow.}
\label{tab:innovations}
\begin{tabular}{|p{3.1cm}|p{5.1cm}|p{5.8cm}|}
\hline
\textbf{Technique / Lesson} & \textbf{Core Idea} & \textbf{Impact / Observation} \\
\hline
Multi-UIP Clause Learning 
& Generate several UIP candidates (e.g., 1-UIP, 2-UIP, last-UIP) and choose by clause-quality metrics (LBD/size). 
& Improves learned-clause quality on some instances; however, computing and scoring multiple UIPs adds overhead, so net wins are instance-dependent and overall gains can be modest relative to the extra cost. \\
\hline
Bandit-Tuned UIP Depth 
& Use a lightweight bandit to select UIP depth or candidate, with reward based on downstream clause utility (e.g., propagation/shortening effects). 
& Reduces the overhead of fixed multi-UIP by focusing effort where it pays off; stabilizes performance across mixed instance families compared with static UIP policies. \\
\hline
Vivification Sensitivity (SAT vs.\ UNSAT) 
& Clause vivification aggressiveness and scheduling (which clauses/when) critically affect outcomes; benefits and costs differ between SAT and UNSAT workloads. 
& Clear trade-off: aggressive vivification can speed UNSAT proofs (stronger clauses) but slow SAT search (overhead, over-tightening); careful thresholds and timing are required to avoid regressions. \\
\hline
Bandit-Tuned Vivification 
& Adapt vivification tiers/effort online using a bandit controller guided by an efficiency signal (e.g., literals removed / clauses checked), optionally with ADAM-like updates. 
& Dynamic vivification adaption; adapts to instance characteristics; reliably recovers most of the UNSAT benefits while avoiding over-vivification on SAT-heavy cases. \\
\hline
Reward Design and Update (Tuning Optimization) 
& (a) Evolve rewards from “variable coverage” proxies toward conflict-/propagation-centered signals that better correlate with PAR-2. (b) Replace simple UCB updates with history-aware and ADAM-like updates for robustness. 
& Conflict- and propagation-aware rewards steer search toward medium–hard instances where PAR-2 is most sensitive; ADAM-like updates improve learning under noisy, non-stationary feedback and reduce flakiness across seeds. \\
\hline
Adaptive Sliding Window (Smoothing) 
& Adjust the statistics window (and/or discounting) based on variance and phase to handle non-stationary rewards. 
& This is proposed to lower variance without incurring high overhead; improves stability during phase transitions (early exploration vs.\ late exploitation) and reduces oscillations in policy. \\
\hline
Multi-Domain Bandit Control 
& Coordinate several subsystems: vivification, restarts (VSIDS/CHB), UIP depth, clause reduction, via concurrent bandits with light cross-regularization. 
& Produces balanced, self-tuned behavior; avoids over-optimizing one component at the expense of others; yields consistent improvements on diverse benchmarks. \\
\hline
BreakID Symmetry Integration 
& Add static symmetry-breaking predicates from BreakID as a pre-/inter-processing step; enable solver to react to highly symmetric CNF structure. 
& Extra constraints introduce overhead, but can substantially shrink symmetric search spaces and unlock additional solves on hard, structured instances; suggests CDCL should adapt to symmetry signatures when detected which lead to other parameters in CDCL and introduced the multi-component tuning mechanism. \\
\hline
Compressed Watch Architecture 
& Compress watch entries to reduce memory footprint and improve cache locality (e.g., short refs for clause and blocker). 
& Significant memory savings and better locality on large formulas; requires careful engineering to avoid correctness pitfalls (e.g., stale refs), but offers clear scalability benefits. \\
\hline
\end{tabular}
\end{table}

The SATLUTION evolution revealed a combination of solver-design insights that extend beyond traditional human-engineered heuristics. These include refinements to clause learning, such as exploring multi-UIP strategies and later enhancing them with bandit-based selection; sensitivity analyses of clause vivification showing trade-offs between SAT and UNSAT cases, followed by adaptive bandit control to balance these effects; improvements in reward design and update mechanisms, shifting from variable-coverage proxies to conflict- and propagation-aware signals with Adam-like updates; adaptive sliding-window schemes to smooth noisy feedback; multi-domain bandit control to coordinate heuristics across solver subsystems; integration of symmetry-breaking (BreakID) \cite{devriendt2016improved} to exploit structural regularities in CNFs; and compressed watch list architectures to reduce memory overhead and improve cache locality. While some ideas remain exploratory, others proved consistently effective and were integrated into later solver generations. A concise overview of these techniques and their observed impact is provided in Table~\ref{tab:innovations}. We note, however, that conducting controlled ablation studies of these learned components remains challenging due to the highly entangled nature of their implementations in a complex SAT solving system. Over successive cycles, the cumulative code modifications expanded the repository by more than 10,000 lines compared to the initial baselines, making it difficult to isolate the effect of any single innovation in isolation.

\bmhead{Acknowledgments}

We thank Dr. Steve Dai and Dr. Brucek Khailany from NVIDIA Research for valuable 
discussions. We are also grateful to Prof.~Zhiru Zhang (Cornell University) 
and Prof.~Alvaro Velasquez (University of Colorado Boulder) for their 
insightful feedback. This research was supported by NVIDIA Research, NVIDIA Academic Grant Program, and in part by the National 
Science Foundation (NSF) under awards \#2349670 and \#2403134.


\bibliography{sn-bibliography}

\begin{appendix}
    \section*{Discussion}

\subsection{SATLUTION flow interactions and limitations}

A central advantage of our framework is the use of Cursor’s rule system, which provides structured scaffolding for planning, programming, and verification, while also enabling efficient token usage through hierarchical token caching and file indexing. These features allowed us to sustain long-horizon repository-scale evolution without overwhelming context windows or incurring prohibitive token costs. However, our experiments also revealed limitations. In fully automated operation—what we refer to as our customized ``YOLO mode``, distinct from the official CLI tool, the agents often struggled, and the flow proved most effective in a semi-automated setup with targeted human intervention. In particular, the agents were prone to failures in SAT/UNSAT correctness checks and deep memory errors such as segmentation faults, where human intervention remained critical to preserve progress. While the planning capabilities of the agents were strong at the level of concrete programming tasks, they lacked sufficient domain-specific knowledge at the idea level, especially for nuanced aspects of SAT solving. To compensate, we incorporated human guidance in later stages by explicitly steering the agent's ``thinking`` process, manually directing higher-level strategies and leaving the lower-level implementation to the agent. This division of labor between rule-guided automation and human-guided conceptual direction proved to be a practical balance, leveraging the strengths of automated coding while mitigating its current weaknesses in domain reasoning.

\subsection{Verifier Plays Critical Role in Self-Evolution Coding Task}

Our overall framework (Fig. \ref{fig:framework}) together with the verifier design (Fig. \ref{fig:verification}) underscores that a carefully calibrated verifier is indispensable for repository-scale solver evolution. In our early experiments, where we relied only on end-of-benchmark validation of SAT/UNSAT outcomes, the evolutionary process was slowed dramatically: errors due to logistics, programming mistakes, or speculative solver behaviors could only be detected after a full evaluation run, wasting at least 1.5 hours per cycle. Such inefficiency not only hindered progress but also destabilized the learning loop, since faulty iterations polluted the reward signal. The introduction of a two-stage verifier fundamentally changed this dynamic. By checking instance-level SAT/UNSAT correctness in Stage 2 and validating UNSAT proofs in Stage 3, the verifier provided immediate, fine-grained signals that prevented wasted evaluations and preserved the integrity of the evolution. Crucially, this verifier was manually engineered prior to the evolutionary process, ensuring domain-specific rigor that agents could not yet provide on their own. Beyond safeguarding correctness, the verifier’s structured feedback also enabled the agent to refine its rule base for planning and programming, embedding verification outcomes directly into the evolutionary logic. Our key takeaway is that, for large-scale and domain-critical tasks like SAT solving, a robust and precisely calibrated verifier is not optional but foundational: it is the anchor that allows agent-driven evolution to proceed efficiently, reliably, and meaningfully.

Looking ahead, we identify verifier construction as a key frontier for future coding agents. At present, human expertise is indispensable in designing robust verification pipelines that are tailored to SAT-domain requirements. However, the ability for agents to autonomously construct, adapt, and optimize their own verifiers would mark a step-change in capability. Such progress would not only accelerate the pace of solver evolution but also extend the framework to new problem domains where correctness guarantees are equally critical. Among these, electronic design automation (EDA) stands out as a particularly impactful application area. Verification and correctness checks are central to large-scale circuit design optimization, spanning logic synthesis, technology mapping, and physical design flows. Errors at these stages can cascade into catastrophic downstream failures, making the availability of rigorous, domain-specific verifiers essential. Coding agents capable of constructing and evolving such verifiers could transform the automation of EDA tasks, enabling not only faster exploration of design spaces but also higher reliability in outcomes. We see the evolution of verifier construction as a milestone that will determine the readiness of coding agents to engage with mission-critical domains like EDA, where correctness is not negotiable but foundational.

\subsection{Runtime feedback hardware setup}

To sustain repository-scale evolution at solver-iteration granularity, we designed a distributed evaluator on an LSF-managed cluster of 800 AMD EPYC 7F72 nodes operating at 3.6 GHz with a three-level cache hierarchy (L1: 32 KB, L2: 512 KB, L3: 16 MB per core). In every evaluation round, two solvers, either successive evolutionary variants or paired A/B test candidates, are launched concurrently, mapped across 400 CNF instances each to saturate all 800 nodes. This configuration ensures full SAT Competition 2024 benchmark coverage per iteration and achieves the theoretical minimum feedback turnaround: 5,000 s, exactly matching the imposed timeout. By aligning runtime parallelism with benchmark granularity, the evaluator attains near-instantaneous fitness assessment at cluster scale, thereby maximizing evolutionary speed and enabling continuous self-adaptation of solvers. As discussed earlier in the Methods section, the evaluator itself was manually implemented prior to the start of the evolution stage, rather than generated by the coding agent. In total, the full set of evolutionary cycles in this work required approximately 90,000 CPU hours of evaluation time.

\subsection{Token usage and code changes}

\begin{figure}[t]
    \centering
\includegraphics[width=0.7\linewidth]{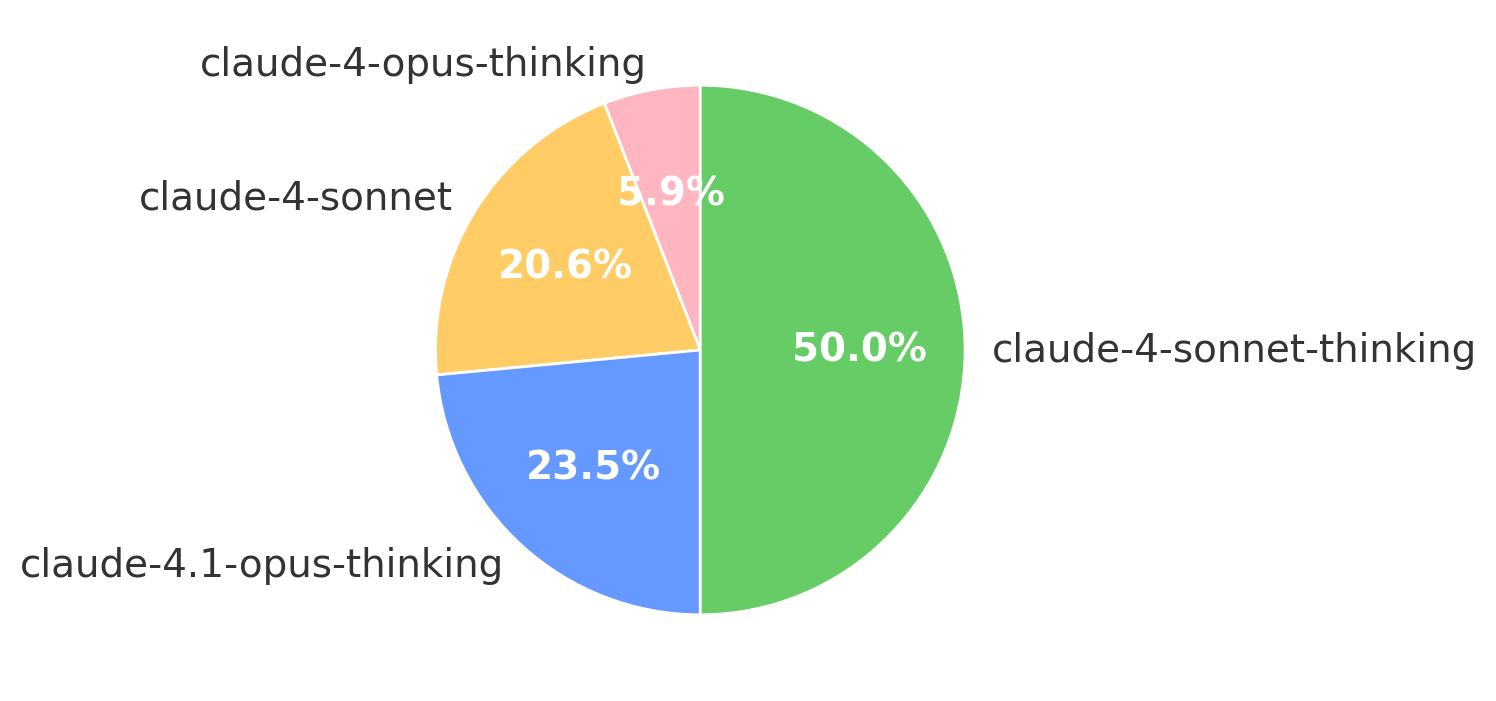}
    \caption{Distribution of model usage in the SATLUTION evolution process.}
    \label{fig:model_usage}
\end{figure}

The token usage analysis reveals a striking asymmetry between different pipeline stages.  
While {cache reads} dominate in raw count, consuming over 314M tokens (more than 93\% of the total), 
the {economic cost} of execution is not directly proportional to raw token counts. 
In current API pricing models, {output tokens are significantly more expensive} than input tokens, 
and among inputs, those processed {without cache write} typically incur the highest cost. 
In our case, the output stage, though representing only $\sim$0.4\% of total tokens, carries a 
disproportionate weight in overall expense. Similarly, the 6.7M tokens of input without cache write, 
despite being only $\sim$2\% of usage, contribute notably to cost. By contrast, the vast volume of 
cache-read tokens, while technically dominant, is priced more cheaply and therefore less critical in 
cost optimization. These results highlight a dual optimization challenge: engineering efforts should 
reduce cache-read redundancy to cut latency and overhead, while cost-aware system design must focus 
on minimizing high-value tokens, especially outputs and uncached inputs, to achieve sustainable 
deployment economics.  The model usage distribution is shown in Figure \ref{fig:model_usage}, illustrating the division of labor among the different Claude models in our framework.

The total lines of code changes related to this project include 12,774 lines of C/C++, 4,292 lines of JSON, 27,593 lines of Markdown, and approximately 20,000 lines of Python. The C/C++ modifications were exclusively devoted to solver implementations. JSON changes largely reflect configuration, structured logging, and intermediate data exchange across evaluation stages. The Markdown content, while the largest in volume, primarily captures self-documentation: recording the history of solver changes, maintaining self-analysis knowledge, and tracking updates to self-evolved rule sets, reward and performance metrics tracking, etc. Python code, by contrast, was generated by the agent for its own targeted analyses, such as feature extraction, additional result profiling, runtime statistics, and verification support. 

\subsection{Total cost Breakdown}

The self-evolution process of SATLUTION incurred two primary categories of cost: token usage in the coding agents and CPU time in distributed evaluation. The total token usage amounted to 1,938.4 USD, dominated by output tokens, which are typically more expensive than input tokens. For runtime evaluation, the system consumed an estimated 90,000 CPU hours. Using AWS on-demand pricing for equivalent compute nodes (ranging from 0.1 to 0.15 USD per CPU hour), this corresponds to an additional 9,000 to 13,500 USD. 

The development of competitive SAT solvers by human experts typically requires months of engineering effort, careful manual tuning of heuristics, and extensive experimental validation. Each solver increment in the SAT Competition ecosystem represents sustained expert labor, often extending over multiple years. By contrast, SATLUTION achieved solver improvements surpassing state-of-the-art competition winners within a matter of weeks, incurring a total cost of less than 20,000 USD across both token usage and compute resources. This is substantially lower than the implicit cost of human time, which, when accounting for multiple full-time researchers and engineers, would amount to several person-months of work.

\newpage

\subsection{Rules and Self-evolve Rules Example}

Examples for self-evolving rulebase are attached. We observed that, for these rules to be most effectively realized by the agent, highly code-oriented instructions (e.g., structured as Bash commands or explicit repository operations) consistently led to stronger agent performance compared to natural language descriptions.



\section*{Supplementary material}

# Rule Compliance Verification System
## ENSURING CURSOR RULES ARE ACTIVELY APPLIED

### **🔍 HOW TO VERIFY RULES ARE BEING FOLLOWED**

## **1. AUTOMATED RULE CHECKING SCRIPTS**

### **Pre-Development Compliance Check:**

```
#!/bin/bash
# rule_compliance_checker.sh SATLUTION_x
set -e

SOLVER=$1
VIOLATIONS=()

echo "🔍 Checking rule compliance for $SOLVER..."

# Check mandatory documentation (Rule 03)
[[ -f "$SOLVER/CHANGELOG.md" ]] || VIOLATIONS+=("Missing CHANGELOG.md")
[[ -f "$SOLVER/HYPOTHESIS.md" ]] || VIOLATIONS+=("Missing HYPOTHESIS.md")
[[ -f "$SOLVER/bin/starexec_run_default" ]] || VIOLATIONS+=("Missing starexec_run_default")

# Check forbidden patterns (Rule 04)
if find "$SOLVER" -name "*.c" -o -name "*.h" | xargs grep -l "malloc\|free\|calloc" | grep -v kissat; then
    VIOLATIONS+=("Forbidden non-Kissat memory functions detected")
fi

if find "$SOLVER" -name "*.c" -o -name "*.h" | xargs grep -l "static.*buffer\|static.*ptr"; then
    VIOLATIONS+=("Forbidden static state detected")
fi

# Check solver branding
if grep -q "SATLUTION_[0-9]*:" "$SOLVER/bin/starexec_run_default" 2>/dev/null; then
    SOLVER_NAME=$(grep "SATLUTION_[0-9]*:" "$SOLVER/bin/starexec_run_default" | head -1 | cut -d: -f1 | tr -d 'c ')
    if [[ "$SOLVER_NAME" != "$SOLVER" ]]; then
        VIOLATIONS+=("Incorrect solver branding: found $SOLVER_NAME, expected $SOLVER")
    fi
else
    VIOLATIONS+=("Missing or incorrect solver branding in starexec_run_default")
fi

# Report violations
if [[ ${#VIOLATIONS[@]} -gt 0 ]]; then
    echo "❌ RULE VIOLATIONS DETECTED:"
    printf ' - %s\n' "${VIOLATIONS[@]}"
    echo "🚨 FIX VIOLATIONS BEFORE PROCEEDING"
    exit 1
else
    echo "✅ All rule compliance checks passed"
fi
```

## **2. CURSOR INTEGRATION VERIFICATION**

### **Check if Cursor is Reading Rules:**

```
# Test if Cursor sees the rules by asking specific questions
echo "Test: Ask Cursor about forbidden patterns - should reference Rule 04"
echo "Test: Ask Cursor about pre-evaluation testing - should reference Rule 01"
echo "Test: Ask Cursor about logging requirements - should reference Rule 03"
```

### **Rule Reference Indicators:**
- **Cursor should mention specific rule numbers** when giving guidance
- **Cursor should reference .cursor/rules/ files** in explanations
- **Cursor should enforce zero-tolerance policies** from forbidden patterns
- **Cursor should require mandatory documentation** before proceeding

## **3. DEVELOPMENT WORKFLOW INTEGRATION**

### **Pre-Commit Hook (Recommended):**
```bash
#!/bin/bash
# .git/hooks/pre-commit
# Automatically check rule compliance before commits

echo "🔍 Running rule compliance check..."
bash .cursor/rules/rule_compliance_checker.sh SATLUTION_current || exit 1
echo "✅ Rule compliance verified"
```

### **Development Session Checklist:**
```markdown
## Before Starting Any Development Session:
```

- [ ] Have I read the latest .cursor/rules/?
- [ ] Am I following Rule 01 (pre-evaluation testing)?
- [ ] Am I documenting changes per Rule 03?
- [ ] Am I avoiding patterns from Rule 04?

## Before Committing Code:
- [ ] Run rule_compliance_checker.sh
- [ ] Verify no forbidden patterns
- [ ] Confirm documentation is complete
- [ ] Check solver branding is correct
```

## **4. CONTINUOUS COMPLIANCE MONITORING**

### **Weekly Rule Audit:**
```bash
#!/bin/bash
# weekly_rule_audit.sh
echo "📋 Weekly Rule Compliance Audit"
echo "Date: $(date)"

# Check all recent SATLUTION directories
for dir in SATLUTION_*; do
    if [[ -d "$dir" ]]; then
        echo "Checking $dir..."
        bash .cursor/rules/rule_compliance_checker.sh "$dir" || echo "⚠️ $dir has violations"
    fi
done

# Check for rule updates needed
echo "🔄 Checking if rules need updates based on recent issues..."
# This could be automated to scan git logs for failure patterns
```

### **Rule Effectiveness Metrics:**
- **Failure Prevention Rate**: Track if rules prevent historical failure patterns
- **Compliance Score**: Percentage of cycles that pass all rule checks
- **Time to Fix Violations**: How quickly rule violations are resolved
- **Rule Coverage**: Whether new failure patterns are covered by existing rules

## **5. VERIFICATION THAT CURSOR IS USING RULES**

### **Test Questions for Cursor:**
1. "What should I check before starting SATLUTION_28?"
   - **Expected**: Reference to Rule 01 (pre-evaluation testing)
2. "Can I use malloc() in my new solver?"
   - **Expected**: Reference to Rule 04 (forbidden patterns)
3. "What documentation do I need for SATLUTION_28?"
   - **Expected**: Reference to Rule 03 (mandatory logging)

### **Cursor Behavior Indicators:**
- **Proactive Rule Enforcement**: Cursor stops you from forbidden patterns
- **Rule-Based Guidance**: Cursor references specific rules in advice
- **Compliance Checking**: Cursor asks about rule compliance
- **Documentation Reminders**: Cursor prompts for required documentation

## **6. RULE UPDATE TRIGGERS**

### **When to Update Rules:**
- **New Failure Pattern Discovered**: Add to Rule 04 (forbidden patterns)
- **Process Improvement Identified**: Update Rule 01 or 03
- **Correctness Issue Found**: Update Rule 02 (critical correctness)
- **Tool/Process Change**: Update all relevant rules

### **Rule Update Process:**
1. **Identify Need**: New failure or process gap
2. **Document Pattern**: Add specific example to appropriate rule
3. **Update Verification**: Add check to compliance scripts
4. **Test Integration**: Verify Cursor follows updated rules
5. **Announce Change**: Log rule update in session documentation

# Automatic Rule Evolution System
## SELF-IMPROVING RULE SYSTEM BASED ON CYCLE LEARNINGS

### **🔄 AUTOMATIC RULE REVISION FOR CONTINUOUS IMPROVEMENT**

## **1. CYCLE-BY-CYCLE RULE EVOLUTION FRAMEWORK**

### **Post-Cycle Rule Analysis Script:**
```bash
#!/bin/bash
# post_cycle_rule_analyzer.sh CYCLE_NUMBER
set -e

CYCLE=$1
ANALYSIS_LOG=".cursor/log/cycle_${CYCLE}_rule_analysis.md"

echo "🔄 Analyzing Cycle $CYCLE for rule updates..."
echo "# Cycle $CYCLE Rule Analysis" > "$ANALYSIS_LOG"
echo "Date: $(date)" >> "$ANALYSIS_LOG"

# Scan for new failure patterns
echo "## 🔍 New Failure Patterns Detected:" >> "$ANALYSIS_LOG"

# Check for new compilation errors
if grep -r "compilation error\|undefined reference\|linking error" SATLUTION_${CYCLE}*/; then
    echo "- **Compilation Issues**: Found new compilation patterns" >> "$ANALYSIS_LOG"
    echo "  - Recommend: Update Rule 01 (pre-evaluation testing)" >> "$ANALYSIS_LOG"
fi

# Check for new correctness failures
if python3 mandatory_test_verifier.py 2>&1 | grep -q "mismatch"; then
    echo "- **Correctness Violations**: New SAT/UNSAT mismatch patterns" >> "$ANALYSIS_LOG"
    echo "  - Recommend: Update Rule 02 (critical correctness)" >> "$ANALYSIS_LOG"
fi

# Check for missing documentation
MISSING_DOCS=()
for solver in SATLUTION_${CYCLE}*; do
    [[ -f "$solver/CHANGELOG.md" ]] || MISSING_DOCS+=("$solver/CHANGELOG.md")
    [[ -f "$solver/HYPOTHESIS.md" ]] || MISSING_DOCS+=("$solver/HYPOTHESIS.md")
    [[ -f "$solver/RESULTS.md" ]] || MISSING_DOCS+=("$solver/RESULTS.md")
done

if [[ ${#MISSING_DOCS[@]} -gt 0 ]]; then
    echo "- **Documentation Gaps**: ${#MISSING_DOCS[@]} missing files" >> "$ANALYSIS_LOG"
    echo "  - Recommend: Strengthen Rule 03 (mandatory logging)" >> "$ANALYSIS_LOG"
fi

# Check for new forbidden patterns
if find SATLUTION_${CYCLE}* -name "*.c" -o -name "*.h" | xargs grep -l "early_termination\|predictive_sat\|heuristic_result" 2>/dev/null; then
    echo "- **New Forbidden Pattern**: Predictive termination variants detected" >> "$ANALYSIS_LOG"
    echo "  - Recommend: Add to Rule 04 (forbidden patterns)" >> "$ANALYSIS_LOG"
fi

echo "✅ Cycle $CYCLE rule analysis complete: $ANALYSIS_LOG"
```

## **2. AUTOMATIC RULE UPDATE MECHANISMS**

### **Rule Update Engine:**
```bash
#!/bin/bash
# auto_rule_updater.sh CYCLE_NUMBER
set -e

CYCLE=$1
ANALYSIS_LOG=".cursor/log/cycle_${CYCLE}_rule_analysis.md"
UPDATE_LOG=".cursor/log/cycle_${CYCLE}_rule_updates.md"

echo "🔄 Updating rules based on Cycle $CYCLE learnings..."
echo "# Cycle $CYCLE Rule Updates" > "$UPDATE_LOG"

# Function to add new forbidden pattern
add_forbidden_pattern() {
    local pattern="$1"
    local explanation="$2"
    local cycle="$3"
    
    # Backup current rule
    cp .cursor/rules/04_forbidden_patterns.md .cursor/rules/04_forbidden_patterns.md.backup
    
    # Add new pattern to Rule 04
    cat >> .cursor/rules/04_forbidden_patterns.md << EOF

### **❌ FORBIDDEN: $pattern (Added Cycle $cycle)**
```c
// LESSON: Cycle $cycle failure
// $explanation

// ❌ FORBIDDEN PATTERN:
$pattern
```

## **3. LEARNING INTEGRATION FRAMEWORK**

### **Automated Learning Pipeline:**
```bash
#!/bin/bash
# cycle_learning_pipeline.sh CYCLE_NUMBER
set -e

CYCLE=$1
echo "🎯 Running complete learning pipeline for Cycle $CYCLE..."

# Step 1: Analyze cycle for new patterns
bash .cursor/rules/post_cycle_rule_analyzer.sh "$CYCLE"

# Step 2: Update rules based on analysis
bash .cursor/rules/auto_rule_updater.sh "$CYCLE"

# Step 3: Validate updated rules
bash .cursor/rules/rule_compliance_checker.sh "SATLUTION_${CYCLE}A" || echo "⚠️ Rule validation needed"

# Step 4: Generate rule evolution summary
echo "📋 Generating rule evolution summary..."
cat > ".cursor/log/cycle_${CYCLE}_rule_evolution_summary.md" << EOF
# Cycle $CYCLE Rule Evolution Summary

## 📊 **Evolution Stats:**
- **Rules Updated**: $(grep -c "✅" .cursor/log/cycle_${CYCLE}_rule_updates.md 2>/dev/null || echo "0")
- **New Patterns Added**: $(grep -c "FORBIDDEN" .cursor/log/cycle_${CYCLE}_rule_updates.md 2>/dev/null || echo "0")
- **Process Improvements**: $(grep -c "pre-evaluation\|logging" .cursor/log/cycle_${CYCLE}_rule_updates.md 2>/dev/null || echo "0")

## 🔄 **Rule System Version:**
- **Previous Version**: Cycle $(($CYCLE - 1))
- **Current Version**: Cycle $CYCLE
- **Total Rules**: $(ls .cursor/rules/*.md | wc -l)

## 📈 **Cumulative Learning:**
- **Total Forbidden Patterns**: $(grep -c "FORBIDDEN:" .cursor/rules/04_forbidden_patterns.md)
- **Total Compliance Checks**: $(grep -c "RULE:" .cursor/rules/01_pre_evaluation_testing.md)
- **Documentation Requirements**: $(grep -c "MANDATORY" .cursor/rules/03_mandatory_logging.md)

## 🎯 **Next Cycle Improvements:**
Based on Cycle $CYCLE learnings, focus areas for Cycle $(($CYCLE + 1)):
- Enhanced pattern detection
- Automated compliance verification
- Improved documentation standards

## **4. CURSOR AGENT INTEGRATION**

### **Meta-Rule for Cursor Agent:**
```
CRITICAL INSTRUCTION FOR CURSOR:
At the END of every cycle (after RESULTS.md is complete),
automatically run the learning pipeline:

1. Execute: bash .cursor/rules/cycle_learning_pipeline.sh CYCLE_NUMBER
2. Review generated rule evolution summary
3. Apply updated rules to ALL future development
4. Reference new patterns in guidance and error prevention
5. Update .cursorrules if major process changes detected

This creates a SELF-IMPROVING rule system that learns from every cycle.
```

### **Automatic Rule Version Management:**
```bash
#!/bin/bash
# rule_version_manager.sh
CURRENT_VERSION=$(date +%Y%m%d)-cycle-$(ls SATLUTION_* | tail -1 | sed 's/SATLUTION_//')

# Backup current rules
mkdir -p .cursor/rules/versions/$CURRENT_VERSION
cp .cursor/rules/*.md .cursor/rules/versions/$CURRENT_VERSION/

# Tag rule version
echo "$CURRENT_VERSION" > .cursor/rules/CURRENT_VERSION
```

```
echo "✅ Rule system versioned: $CURRENT_VERSION"
```

## **5. INTEGRATION WITH .CURSORRULES**

### **Addition to Main .cursorrules File:**
```markdown
## AUTOMATIC RULE EVOLUTION (Meta-Rule)

### **Post-Cycle Learning Protocol:**
After every cycle completion (when RESULTS.md is written):
1. **MANDATORY**: Run cycle learning pipeline
2. **AUTOMATIC**: Update rules based on new failure patterns
3. **INTEGRATION**: Apply updated rules to all future guidance
4. **VERIFICATION**: Test rule compliance on next solver

### **Continuous Improvement Cycle:**
```
New Failure → Pattern Analysis → Rule Update → Cursor Integration → Prevention
     ↑                                                                    ↓
     Learning ←————————————————————————————————————————————————————————
   Application
```

### **Rule Evolution Triggers:**
- **Immediate**: Critical correctness failures (update Rule 02)
- **Post-Cycle**: Performance issues or process gaps (update Rules 01, 03)
- **Weekly**: Pattern accumulation analysis (update Rule 04)
- **Monthly**: Rule system effectiveness review (update Rule 05)
```

## **6. IMPLEMENTATION CHECKLIST**

### **To Enable Automatic Rule Evolution:**
- [ ] ✅ Create rule analysis scripts (post_cycle_rule_analyzer.sh)
- [ ] ✅ Create rule update engine (auto_rule_updater.sh)
- [ ] ✅ Create learning pipeline (cycle_learning_pipeline.sh)
- [ ] ✅ Create version management (rule_version_manager.sh)
- [ ] ⭕ Add meta-rule to .cursorrules (NEXT STEP)
- [ ] ⭕ Test on Cycle 27 completion (VALIDATION)

### **Expected Benefits:**
- **Reduced Repeated Failures**: Patterns caught automatically
- **Faster Issue Resolution**: Rules updated immediately after learning
- **Cumulative Intelligence**: Each cycle makes system smarter
- **Zero Manual Rule Updates**: Cursor learns and adapts automatically

### **Success Metrics:**
- **Pattern Prevention Rate**: % of historical patterns prevented
- **Rule Coverage**: % of failure modes covered by rules
- **Adaptation Speed**: Time from failure to rule update
- **Cursor Compliance**: % of rule-guided development decisions

## **7. USAGE EXAMPLE**

### **End of Cycle 27 (Automatic Sequence):**
```bash
# 1. Complete evaluation
python3 post_evaluation_analyzer.py SATLUTION_26A.csv SATLUTION_27A.csv

# 2. Write RESULTS.md for both variants
# ... [manual documentation] ...

# 3. AUTOMATIC RULE LEARNING (Cursor triggered)
bash .cursor/rules/cycle_learning_pipeline.sh 27

# 4. RESULT: Updated rules for Cycle 28
# - New forbidden patterns from any Cycle 27 issues
# - Enhanced pre-evaluation tests if needed
# - Improved documentation requirements
# - Cursor automatically applies new rules to Cycle 28 guidance

# 5. Cycle 28 development begins with evolved rule system
```

This creates a **self-improving AI development system** where both the SAT solvers
AND the development rules evolve together through principled learning.


\end{appendix}
\end{document}